\documentclass{article}
\usepackage{PRIMEarxiv}
\pdfoutput=1
\usepackage[utf8]{inputenc} 
\usepackage[T1]{fontenc}    
\usepackage{hyperref}       
\usepackage{url}            
\usepackage{booktabs}       
\usepackage{amsfonts}       
\usepackage{nicefrac}       
\usepackage{microtype}      
\usepackage{lipsum}
\usepackage{fancyhdr}       
\usepackage{graphicx}       
\graphicspath{{media/}}     

\usepackage{amssymb}
\usepackage{setspace}
\usepackage[figuresright]{rotating}
\usepackage[english]{babel}
\usepackage[utf8]{inputenc}
\usepackage{amsmath}
\usepackage[numbers]{natbib}
\usepackage{graphicx}
\usepackage{caption}
\usepackage{subcaption}
\usepackage{amsfonts}
\usepackage{verbatim}
\usepackage{algpseudocode}
\usepackage{algorithm}
\usepackage[colorinlistoftodos]{todonotes}
\usepackage{graphicx}
\usepackage{pythonhighlight}
\usepackage{listings}

\newcommand{\be}{\begin{equation}}
\newcommand{\ee}{\end{equation}}
\usepackage{lineno}

\title{An exact mapping from ReLU networks to \\spiking neural networks}
\author{
  \textbf{Ana Stanojevic}$^{1, 2}$ \hspace{1cm}  \textbf{Stanis{\l}aw Wo{\'z}niak}$^{1}$  \hspace{1cm} \textbf{Guillaume Bellec}$^{2}$  \\  \textbf{Giovanni Cherubini}$^{1}$ \hspace{1cm} \textbf{Angeliki Pantazi}$^{1}$  \hspace{1cm} \textbf{Wulfram Gerstner}$^{2}$   \\
   ${^{1}}$ {IBM Research Europe – Zurich, R{\"u}schlikon, Switzerland} \\
  ${^{2}}$ {\'E}cole polytechnique f{\'e}d{\'e}rale de Lausanne, Lausanne EPFL, Switzerland \\
}

\begin{document}
\maketitle

\begin{abstract}
Deep spiking neural networks (SNNs) offer the promise of low-power artificial intelligence. However, training deep SNNs from scratch or converting deep artificial neural networks to SNNs without loss of performance has been a challenge.
Here we propose an exact mapping from a network with Rectified Linear Units (ReLUs) to an SNN that fires exactly one spike per neuron. For our constructive proof, we assume that an arbitrary multi-layer ReLU network with or without convolutional layers, batch normalization and max pooling layers was trained to high performance on some training set. Furthermore, we assume that we have access to a representative example of input  data used during training and to the exact parameters (weights and biases) of the trained ReLU network. The mapping from deep ReLU networks to SNNs causes zero percent drop in accuracy on CIFAR10, CIFAR100 and the ImageNet-like data sets Places365 and PASS. More generally our work shows that an arbitrary deep ReLU network can be replaced by  an energy-efficient single-spike neural network without any loss of performance.  
\end{abstract}


\section{Introduction}
Energy consumption of deep artificial neural networks (ANNs) with thousands of neurons poses a problem not only during training \cite{energy_server}, but also during inference \cite{gpt3}. Among other alternatives \cite{binarized,mobilenets, efficientnet}, 
hardware implementations of spiking neural networks (SNNs) 
\cite{neuromorphic_in_memory1, neuromorphic_in_memory2, Goltz-Petrovici21,gallego2020event,Goltz21} have been proposed  as an energy-efficient solution, not only for large centralized applications, but also for computing in edge devices \cite{dl_edge_merge, google_nn, ai_on_edge}.
In SNNs neurons communicate by ultra-short pulses, called action potentials or spikes, that can be considered as point-like events in continuous time. In deep multi-layer SNNs, if a neuron in layer $n$ fires a spike, this event causes a change in the voltage trajectory of neurons in layer $n+1$. If, after some time,  the trajectory of a neuron in layer $n+1$ reaches a threshold value, then this neuron fires a spike. 

 While there is no general consensus on how to best decode spike trains in biology \cite{Rieke96,Gerstner02,Pillow08},
multiple pieces of evidence indicate that immediately after an onset of a stimulus, populations of neurons in auditory, visual, or tactile sensory areas respond in such a way that the timing of the first spike of each neuron after stimulus onset contains a high amount of information about the stimulus features
\cite{temporal_retina,temporal_tactile,temporal_owls}. These and similar observations have triggered the idea that, immediately after stimulus onset,  an initial wave of activity  is triggered and travels across several brain areas in the  sensory processing stream \cite{Optican87,Thorpe96,Thorpe01,Hung05,Yamins16}.
We take inspiration from these observations and assume in this paper that information is encoded in the exact spike times of each neuron and that spikes are transmitted in a wave-like manner across layers of a deep feedforward neural network.

Specifically, we use coding by time-to-first-spike (TTFS) \cite{Gerstner02}, a timing-based code originally proposed in neuroscience
\cite{Gerstner02,temporal_retina,temporal_tactile,Thorpe01}, which has recently attracted substantial attention in the context of neuromorphic implementations
\cite{Goltz-Petrovici21,gallego2020event,Goltz21, Rueckauer18,comsa, mostafa, Zhang21, Kheradpishe20, stanojevic}. 
In our implementation of  TTFS coding, each neuron fires exactly one spike. The stronger the input to a given neuron the earlier it fires. 
Coding schemes with at most one spike per neuron are intrinsically sparse in terms of the number of spikes. Since spike generation is a costly process from the energetic point of view, TTFS coding paves the way towards implementations with low energy consumption.  
 
While a relation between ReLU networks and networks of non-leaky integrate-and-fire neurons with TTFS coding has been suggested before \cite{Rueckauer18, Zhang21, Kheradpishe20}, there has been so far a major obstacle that prevented a successful exact mapping from an abitrary deep ANN to a deep SNN with TTFS coding. In a standard ANN, time is discretized and layers are processed one after the other. The hard problem of an exact conversion of ReLU activations into spike times arises from the fact that in an SNN spikes are point-like events that arrive asynchronously in continuous time. Imagine a neuron in a ReLU network that receives several positive inputs that add up to a value of 0.8 and several negative inputs that add up to a value of -1.0.  Assuming a vanishing bias, the output of the unit is zero. However, if in the corresponding neuron of the SNN all the positive inputs have arrived {\em before} the negative inputs, the spiking neuron will have emitted a spike as soon as the firing threshold is reached \cite{Rueckauer18}. Yet, it is impossible to ``call back'' the spike later on so as to cancel it. A potential solution to this problem is to consider that each layer starts its computation only once the calculation in the previous layer has finished. In a recent conversion scheme a time-dependent threshold was  used to enforce the necessary waiting time \cite{Rueckauer18}. However, due to imperfect conversion, there exists a small, but not negligible, loss in the final performance measure. Other conversion approaches for deep networks use custom activation functions \cite{Bu22} for ANN training, or numerically optimized multi-spike codes for a given activation function \cite{Stockl21}. 
Moreover, most  earlier approaches assumed that weights and biases of the ReLU network can be used as such in the corresponding SNN, but there is no fundamental reason why this should be the case. 

What we would like for an exact mapping from ANN to SNN with TTFS coding  is (i) a guarantee that no neuron fires too early so as to avoid the hard problem mentioned above; and (ii) a coding rule of how to translate the output  $\bar{x}_i^{(n)}$ of neuron $i$ in layer $n$ of the ANN into a spike time of a corresponding neuron in the SNN. 
As an aside, we note that the the hard problem of TTFS coding disappears if the trajectory of spiking neurons is always positive.
The ideal mapping approach starts from a standard ReLU network with or without convolutional layers, batch normalization, and max pooling and maps it by a potentially nonlinear transformation of parameters to a corresponding SNN without any performance loss using a well-defined mapping rule from rates to spike timings.

\par In this paper we construct an explicit  mapping that addresses the points above and 
guarantees the mathematical equivalence of an SNN with the corresponding ANN. We assume that there is a pretrained ReLU network and that we have access to its weights and biases as well as the input data on which the network was trained. Using TTFS coding, we propose a lossless conversion which maps a deep ANN with ReLUs to an equivalent deep SNN with non-leaky integrate-and-fire units. In contrast to other methods which require fine-tuning of the SNN or training an SNN from scratch \cite{Goltz-Petrovici21, Kheradpishe20, Neftci19, Zenke18, Bohte2002, Tavanaei19, Zenke21, learning_gd_bio_bellec,learning_gd_snu, yan_temporal}, the goal of our work is to derive the final SNN from the pretrained deep ReLU network. Further numerical approximation or optimization steps \cite{Rueckauer18,Stockl21,Bu22}  are not needed in our approach. The key to building an efficient TTFS conversion method is to derive an exact mathematical equivalence between an arbitrary deep ReLU network and the corresponding spiking network.
 Using standard pretrained models available online, we demonstrate 0\% conversion loss on CIFAR10 and CIFAR100 \cite{cifar10_100} datasets as well as larger ImageNet-like \cite{imagenet, imagenet_privacy} datasets such as Places365 \cite{places365} and PASS \cite{PASS} without any training or fine-tuning.

\section {Results}

The subsection 'Main Theoretical Result' formulates the precise claim of a family of exact mappings from ANN to SNN. We then present the main ideas of the proof for one specific mapping scheme before we sketch alternative mapping schemes. Finally we test our mapping algorithm on benchmark datasets.

\subsection{Main Theoretical Result}

{\bf Definition: Deep ReLU Network}.
{\em A Deep ReLU Network consists of $M\ge 1$ layers of hidden neurons with full or convolutional feedforward connectivity. Each neuron implements a piecewise linear function $x_i^{(n)} = [a_i^{(n)}]_+$, where $[\,]_+$ denotes rectification, and its activation variable $a_i^{(n)}$ is defined as:
\be\label{eq-ReLU}
a_i^{(n)}
= \sum_j w_{ij}^{(n)}
x_j^{(n-1)} + b_i^{(n)}
\ee
with weights $w_{ij}^{(n)}$ and a bias $b_i^{(n)}$. An upper index $n=0$ refers to the input layer and the i-th input is denoted with $x_i^{(0)}$. Optionally the network may also contain processing steps of max pooling and batch normalization.} 

\begin{figure*}[!t]
\centering
\includegraphics{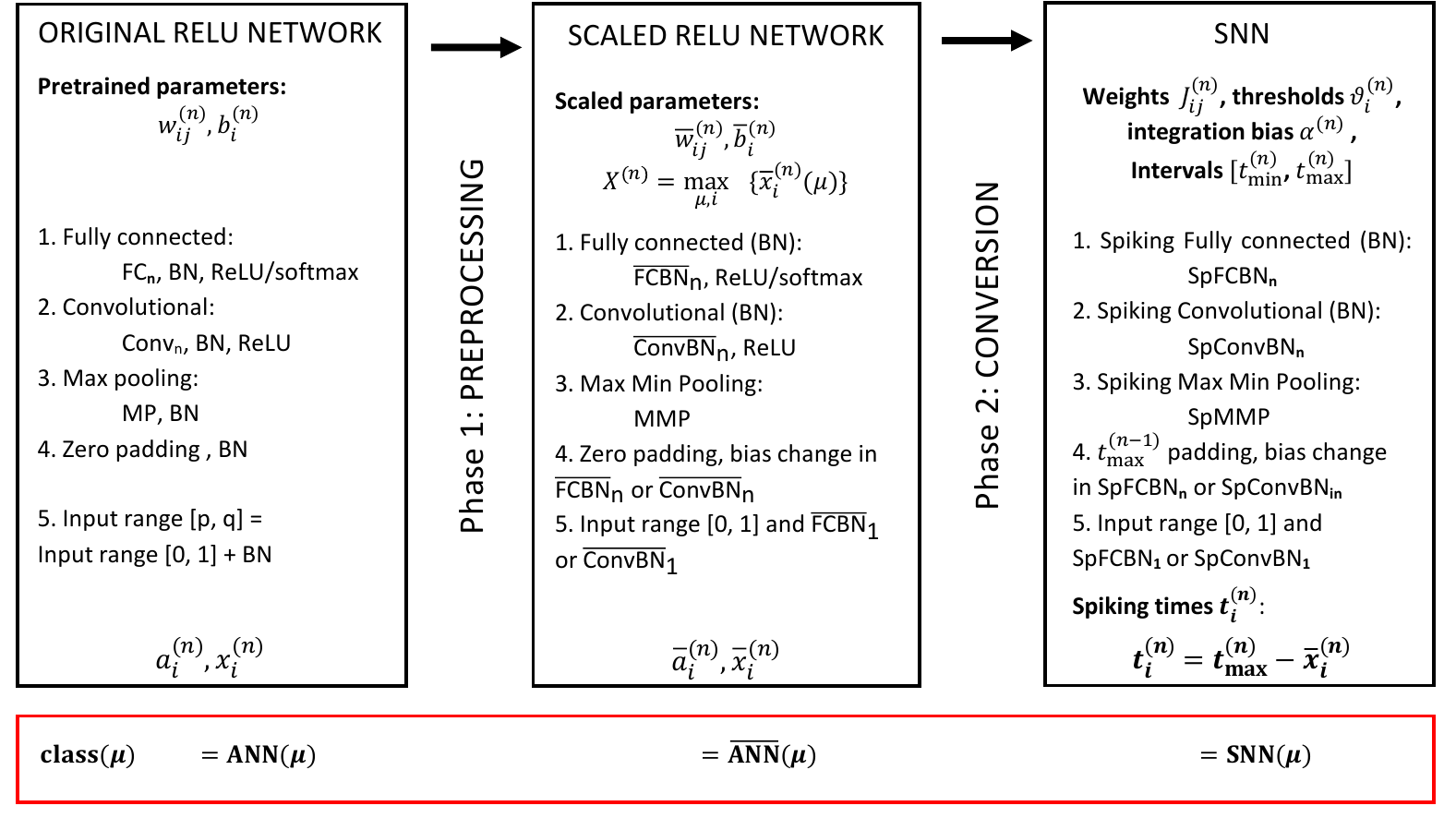}
\caption{
\begin{small}
Two phases of constructing a bidirectional mapping from ANN to SNN. In the first phase, Fully Connected (FC$\mathrm{_n}$) or Convolutional (Conv$\mathrm{_n}$) layers are identified; batch normalization (BN) is fused with the neighbouring layers yielding layers $\overline{\rm FCBN}_\mathrm{n}$ and $\overline{\rm ConvBN}\mathrm{_n}$, respectively; in case of zero padding, biases are adjusted at certain locations; max pooling steps are transformed to combined 'Max Min Pooling' (MMP) steps; and inputs are normalized to $[0,1]$. Then the scaling symmetry of ReLU $x_i^{(n)} = [a_i^{(n)}]_+  = C [a_i^{(n)}/C]_+$ for an arbitrary constant $C>0$ is applied to bring the weights of the ReLU network into a desired range and the maximum output $X^{(n)}$ for each layer is calculated. Overline indicates scaled ReLU network. In the second phase, the resulting  parameters $\{\bar{w}_{ij}^{(n)}, \bar{b}_i^{(n)}\}$ are mapped to the parameters $\{J_{ij}^{(n)}, \vartheta_i^{(n)}, \alpha^{(n)}, t_{\mathrm{min}}^{(n)}, t_{\mathrm{max}}^{(n)}\}$ of the SNN. For an arbitrary input data point  $\mu$ the three networks have the same values in the output layer (before applying softmax) and are therefore predicting the same $\mathrm{class}(\mu)$.  
\end{small}
}
\label{fig:Fig0}
\end{figure*}

\par Our aim is to map each neuron of the Deep ReLU network to an integrate-and-fire neuron in the SNN so that each neuron fires exactly once. A spike at time $t_j^{(n-1)} $ of neuron $j$ in layer $n-1$ generates a step current input with amplitude $J_{ij}^{(n)}$ into neuron $i$ of layer $n$.
The voltage trajectory of neuron $i$ in layer $n$ evolves according to
\be\label{voltage-main}
\frac{\mathrm{d}V_i^{(n)}}{\mathrm{d}t} = \alpha_i^{(n)}\, H(t-t_{\mathrm{min}}^{(n-1)})
  + \sum_j
  J_{ij}^{(n)}
  H (t- t_j^{(n-1)} ) + I_i^{(n)}(t)
\ee
where $H$ denotes the Heaviside step function with $H(x)=1$ for $x>0$
or zero otherwise.
The integration  starts at time $t_{\mathrm{min}}^{(n-1)}$. 
The slope parameters $\alpha_i^{(n)}$, the weights $J_{ij}^{(n)}$, and the thresholds $\vartheta_i^{(n)}$  are parameters of the SNN. Neuron $i$ in layer $n$ may also receive an additional input $I_i^{(n)}(t)$. If the trajectory of $V_i^{(n)}$ crosses the threshold $\vartheta_i^{(n)}$ at time $t$, then $t=t_i^{(n)}$ is the firing time of neuron $i$ in layer $n$. In our mapping, we use $I_i^{(n)}(t)$ to induce a short current pulse so as to trigger a spike at time 
$t_{\mathrm{max}}^{(n-1)}$ if neuron $i$ has not fired before. 
 
We claim that any Deep ReLU network can be mapped exactly to an SNN with integrate-and-fire neurons. 

\vspace{10mm}

{\bf Theorem: Exact mapping from ANN to SNN}.
{\em Given the network parameters $\{w_{ij}^{(n)}, b_i^{(n)}\}$ of a Deep ReLU network that has been trained to high performance on a training set and given access to a representative subset of the input data of the training set, there exists a family of lossless  bidirectional mappings from the Deep ReLU network to an SNN with TTFS coding where each ReLU is replaced by an integrate-and-fire unit with dynamics as in Eq. (\ref{voltage-main}) and parameters $\{J_{ij}^{(n)}, \vartheta_i^{(n)}, \alpha_i^{(n)}, t_{\mathrm{min}}^{(n)}, t_{\mathrm{max}}^{(n)}\}$.
}

\vspace{1mm}

{\bf Remarks}. (i) The theorem mentions a family of mappings since the mapping is not unique, i.e., different combinations of parameters in the SNN give rise to an exact mapping. (ii) In the family of mappings that we consider  each neuron emits at most a single spike. (iii) A consequence of the exact mapping is that both SNN and ANN have exactly the same performance on a sample-by-sample basis: if for a specific sample the prediction of the ANN is wrong then this is also the case for the SNN, and vice versa. (iv) One of the potential mappings is such that slope parameter $\alpha_i^{(n)}$ is identical for all neurons and all layers as stated in the following corollary. 

\vspace{1mm}

{\bf Corollary: Mapping with fixed $\alpha$}. {\em An Exact mapping from ANN to SNN is possible with a slope parameter
$\alpha_i^{(n)} = \alpha>0$ that is identical for all neurons in all layers. Moreover, we may choose $t_{\mathrm{min}}^{(n)}=t_{\mathrm{max}}^{(n-1)}$}.

\begin{figure*}[!t]
\centering
\includegraphics{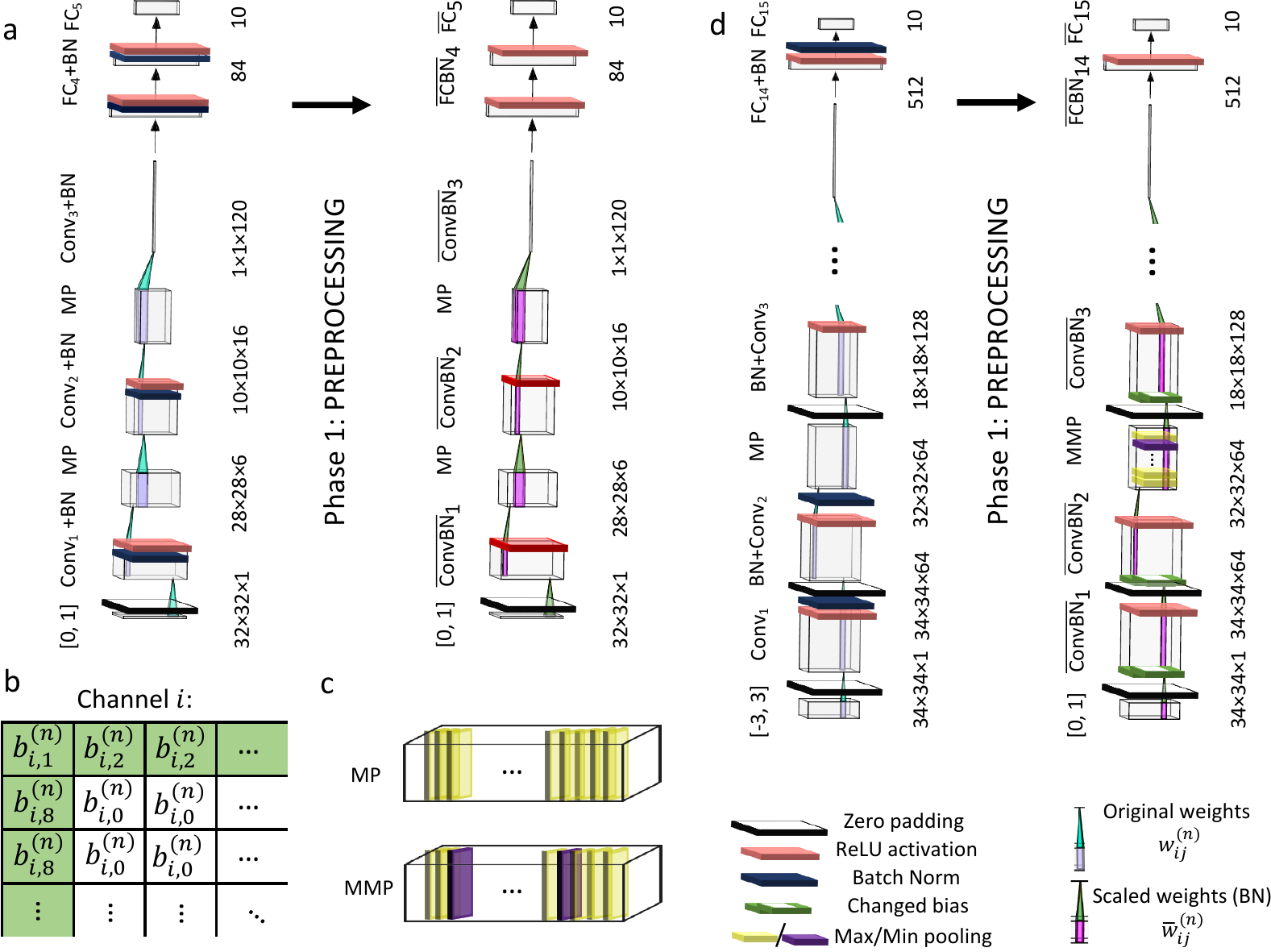}
\caption{
\begin{small}
\textbf{Preprocessing} for LeNet5 and VGG16 networks: \textbf{a.} LeNet5 original ReLU network with batch normalization (BN, black rectangular sheet) before the activation function (red sheet); during preprocessing batch normalization is fused with {\em previous} convolutional (Conv$\mathrm{_n}$) and fully connected (FC$\mathrm{_n}$) layers.
\textbf{b.} When fusing a batch normalization layer with the next convolutional layer containing zero padding, some of the biases (in green) are changed; \textbf{c.} When fusing batch normalization with the next convolutonal layer with max pooling (yellow) in between, specific channels might be changed to use min pooling function (violet). \textbf{d.} VGG16 original ReLU network used for CIFAR10 classification with batch normalization after the activation function; during preprocessing batch normalization is fused with {\em following} convolutional and fully connected layers. 
\end{small}
} 
\label{fig:Fig1}
\end{figure*}

\subsection{Proof Sketch of Main Theoretical Result}
Our proof is constructive, i.e., we propose an explicit mapping algorithm. The arguments work for arbitrary $\alpha_i^{(n)}$. At the end of the argument we set  $\alpha_i^{(n)}=\alpha$ to instantiate the conditions of  the Corollary; see Methods for details.  
The algorithm has two phases, see Fig. \ref{fig:Fig0}.

{\bf Phase 1: Preprocessing}. The original ReLU network undergoes lossless preprocessing such that the network expects input in the $[0,1]$ range; furthermore batch normalization steps are removed by fusing them with the weights of neighbouring layers; see Fig. \ref{fig:Fig1}, Algorithm \ref{alg:preproc} and Methods for details. 

Importantly, and different to other studies in the field of network conversion, we use the known scaling symmetry of the ReLU activation function, i.e., $[a_i^{(n)}]_+  = C [a_i^{(n)}/C]_+$ for an arbritrary constant $C >0$, to implement a nonlinear transformation from the original weight and bias parameters of the ReLU network to new  parameters $\{ \bar{w}_{ij}^{(n)}, \bar{b}_i^{(n)} \}$. After the transformation we can guarantee that  the sum of the weights in each neuron is bounded in the range
$ -B_{\mathrm{low}} \le \sum_j \bar{w}_{ij}^{(n)} \le 1-\delta$ for hyperparameters $B_{\mathrm{low}}>0$ and $0<\delta\ll 1 $.  The transformation  proceeds layer-wise from the input to the output layer and does not change the network output. Neuron $i$ in layer $n$ of the scaled ReLU network  has 
an activation variable  $\bar{a}_i^{(n)}$ and output $\bar{x}_i^{(n)}$.  The mapping is bidirectional so given the quantities $\bar{a}_i^{(n)}$ and $\bar{x}_i^{(n)}$ of the scaled network we can recover the variables $a_i^{(n)}$ and $x_i^{(n)}$ of the original network; see Methods Eqs. (\ref{eq:a_bar_pos})-(\ref{eq:x_bar_neg}). Finally, since we have access to a representative sample of input data used during training, we extract $X^{(n)}$, the maximal activation of the rescaled ReLUs in layer $n$, across the input data  and all neurons in layer $n$. 

\begin{figure*}[!t]
\centering
\includegraphics[width=0.6\textwidth]{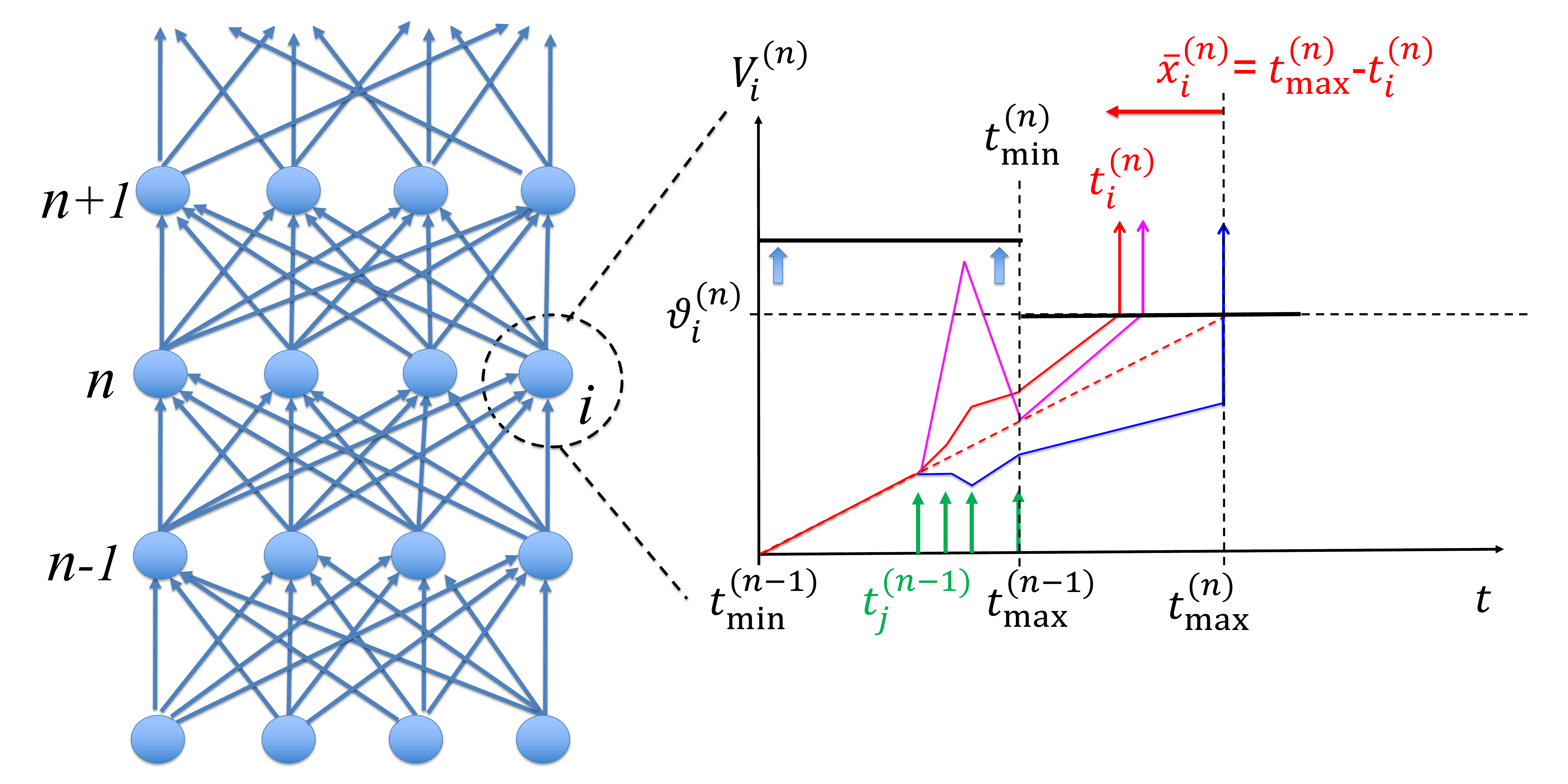}
\caption{
\begin{small}
\textbf{Mapping} to multi-layer SNN with TTFS encoding and integrate-and-fire units. \textbf{Left:} Neurons
  in layer $n-1$ are connected to neuron $i$ in layer $n$.
  \textbf{Right:} Spikes from neurons $1\le j
  \le N^{(n-1)}$
arrive at times $t_j^{(n-1)}$
    (green vertical arrows).
  The red trajectory shows the evolution of the potential $V_i^{(n)}$
  as a function of time. Neuron $i$ fires at time 
  $t_i^{(n)}$ (red vertical arrow) when $V_i^{(n)}$ reaches the threshold $\vartheta_i^{(n)}$.
    The output value 
  $\bar{x}_i^{(n)}$ of the corresponding ReLU corresponds to the time difference between
    $t_i^{(n)}$ and $t_{\mathrm{max}}^{(n)}$.
    Other neurons in layer $n$ fire at other moments (blue, magenta).
    No neuron in layer $n$ can fire later than $t_{\mathrm{max}}^{(n)}$. Our exact mapping procedure guarantees, that for all neurons the slope of the trajectory at the moment of spike firing is positive. Moreover, since in our theorem the threshold is arbitrarily high  (thick blue vertical arrows) for $t<t_{\mathrm{min}}^{(n)}$, all firings in layer $n$ occur for $t>t_{\mathrm{min}}^{(n)}$ so that the hard problem of late inhibitory input is solved.
    \end{small}
    }
\label{fig:Fig2}
\end{figure*}

{\bf Phase 2: Conversion}.
To construct an exact loss-free conversion of the scaled ReLU network  to the network of spiking neurons we exploit six essential ideas (see Methods for details):

(i) {\em Choice of TTFS code}.
We construct a mapping such that each neuron $i$ in layer $n$ of the SNN emits exactly one spike
at $t_i^{(n)}$, where $t_{\mathrm{min}}^{(n)} < t_i^{(n)} \le t_{\mathrm{max}}^{(n)} $ (see Fig. \ref{fig:Fig2}). Positive activation leading to a ReLU output $\bar{x}_i^{(n)}=\bar{a}_i^{(n)}>0$
corresponds to an {\em early}  firing time
$t_i^{(n)}= t_{\mathrm{max}}^{(n)}-\bar{a}_i^{(n)}$, or equivalently,
$\bar{a}_i^{(n)} =t_{\mathrm{max}}^{(n)} - t_i^{(n)}$. 
Thus, spike times depend {\em linearly} on the output $\bar{x}_i^{(n)}$ of active  ReLU neurons. 
Moreover
if a neuron in layer $n$ has not fired before time $t_{\mathrm{max}}^{(n)} $, it receives an additional external input pulse $I_i^{(n)}(t) = R \,\vartheta_i^{(n)} \, \delta(t-t_{\mathrm{max}}^{(n)})$ with $R\gg 1$ that triggers immediate firing at time $t_{\mathrm{max}}^{(n)}$. The parameters $\alpha_i^{(n)}$, $\vartheta_i^{(n)}$, the times $t_{\mathrm{min}}^{(n)} = t_{\mathrm{max}}^{(n-1)}$, as well as the weights $J_{ij}^{(n)}$  of the spiking network are determined during the conversion for all $i,j,n$  as described in Methods (see Algorithm \ref{alg:mapping}) and are kept fixed thereafter.
With this coding scheme each  neuron in layer $n-1$ fires exactly once up to $t_{\mathrm{max}}^{(n-1)}= t_{\mathrm{min}}^{(n)}$. Therefore for $t>t_{\mathrm{min}}^{(n)}$ all input spikes to neurons in layer $n$ have arrived.

(ii) {\em Slope of trajectory}. 
Since for $t>t_{\mathrm{min}}^{(n)}$ all input spikes to neurons in layer $n$ have already arrived,
the  trajectory of  neuron $i$ in layer $n$ has,
a {\em constant} slope
$\sum_j   J_{ij}^{(n)} + \alpha^{(n)}$ which is {\em independent} of the sequence of spike arrivals; see. Eq. (\ref{voltage-main}). This slope is positive thanks to  the rescaling  of the ReLU  weights during the preprocessing phase.

(iii) {\em Weight conversion}. Since for $t>t_{\mathrm{min}}^{(n)}$  the trajectories have positive slope, the mapping from activations in the ReLU to firing times in the SNN  can be derived from the threshold-crossing condition $V_i^{(n)} = \vartheta_i^{(n)}$ for each neuron $i$ in layer $n$.  Evaluating this condition 
 yields the nonlinear conversion of weights
\be\label{eq-J-mapping}
J_{ij}^{(n)} = \frac{\alpha_i^{(n)}}{1-\sum_j \bar{w}_{ij}^{(n)}} \,
\bar{w}_{ij}^{(n)} \
\ee
which is invertible. 
A similar invertible relation holds for the bias parameter (see Methods). Thus weights in the scaled ANN can be mapped to weights in the SNN  without sign change. Summation over $j$ on both sides of Eq. (\ref{eq-J-mapping}) shows that the slope has a value $ \alpha_i^{(n) }+ \sum_j   J_{ij}^{(n)}=
\alpha_i^{(n)}/(1-\sum_j   \bar{w}_{ij}^{(n)} ) >0 $.
Thus, once all input spikes have arrived, the  slope of the trajectories is positive because of the weight rescaling 
$\sum_j \bar{w}_{ij}^{(n)}<1$ as claimed in point (ii). This is the key motivation for the weight rescaling in Phase 1.

(iv) {\em Choice of $t^{(n)}_{\mathrm{max}}$ }. Given our TTFS code,  we know that a stronger activation leads to earlier spikes, yet we have to make sure that no neuron in layer $n$ fires before the last spike of neurons in layer $n-1$. The earliest possible spike  in layer $n$ 
occurs at time $t^{(n)}_{\mathrm{max}} - X^{(n)}$ where $X^{(n)}$ is the maximal activation of ReLU neurons in layer $n$ identified during the preprocessing phase. 
We therefore set $t^{(n)}_{\mathrm{max}} = t^{(n)}_{\mathrm{min}} + (1 + \zeta) X^{(n)}$, where $\zeta>0$. In practice (see below) a value of $\zeta=0.5$ works well.

(v) {\em Choice of threshold}. By definition of our TTFS code,  $t^{(n)}_{\mathrm{max}}$  is the time when a neuron in layer $n$ that corresponds to a  ReLU with activation $\bar{a}_i^{(n)}=0$  reaches the threshold  $\vartheta_i^{(n)}$; therefore this condition defines the value of the  threshold.  Because of different biases and different weights for different neurons, 
the thresholds $\vartheta_i^{(n)}$ are neuron-specific (see Methods).  {\em This finishes the proof in the general case.}

 (vi) {\em Free slope parameter}. Since the slope factor  $\alpha_i^{(n)}$ is a free parameter,  we can arbitrarily set $\alpha_i^{(n)} =1$ for all neurons across all layers $1\le n \le M$. This yields the Corollary. The condition of the Corollary is the specific case used in the simulations.
 \vspace{5mm}
\vspace{1mm}

{\bf Remark}. We may wonder how the above points solve the hard problem of TTFS coding.
Our analysis above makes
 no statement about the trajectories of neurons in layer $n$ for the time $ t< t^{(n)}_{\mathrm{min}}$. Therefore 
we formally  set  the threshold for   $t<t^{(n)}_{\mathrm{min}}$ to an arbitrarily high value to ensure that no spike occurs before $t^{(n)}_{\mathrm{min}}$.
As mentioned  under point (ii),  our method guarantees a  positive slope {\em after 
$t^{(n-1)}_{\mathrm{max}} = t^{(n)}_{\mathrm{min}}$.  }
Since during the allowed spiking interval $[t^{(n)}_{\mathrm{min}},
t^{(n)}_{\mathrm{max}}]$ the slope is fixed and positive, a spike never 
 needs to be "called back". Because of  the preprocessing, we know that 
  $\sum_j\bar{w}_{ij}^{(n)}>-B_{\mathrm{low}}$. For example, a choice $B_{\mathrm{low}}=10$  and $\alpha^{(n)} =1$ yields a slope larger than 1/11. Furthermore, because of our choice of 
$t^{(n)}_{\mathrm{max}}$ under point (iv) 
we know that the interval 
$[   t^{(n)}_{\mathrm{min}},  t^{(n)}_{\mathrm{max}}]$ is long enough to 
allow 
even the most activated neuron to fire at the correct time. Finally, because of our choice of TTFS code under point (i) we are sure that all neurons in layer $n-1$ have fired before or at  $t^{(n)}_{\mathrm{min}}$.  These choices together solve the hard problem of TTFS coding.

\subsection{Examples of equivalent mappings}

As stated in the main Theorem, the mapping from ANN to SNN is not unique; rather there is a family of equivalent mappings. Here we present several concrete implementation schemes.

\subsubsection{Mapping with guaranteed positive slope}

In the proof sketch above it was shown that the slope of all neurons is always positive once all input spikes have been received. However, we cannot exclude that {\em before} the time $t_{\rm min}^{(n)}$ the trajectory transiently has a negative slope; see Fig. \ref{fig:Fig2}. If this is desired for some application, we 
can use the free parameter $\alpha_i^{(n)}=\alpha^{(n)}$ to ensure that the slope of the trajectory is {\em always} non-negative, even before $t_{\rm min}^{(n)}$.
To do so, we sum over all negative weights incoming to a given neuron and choose the slope parameter in layer $n$ such that 
\be\label{eq-stricter}
\alpha^{(n)} + \min_i \sum_j  
J_{ij}^{(n)} H(-J_{ij}^{(n)}) > 0
\ee
This ensures that the slope is positive not only if all inhibitory spikes arrive before the first excitatory spike, but also for all other possible timings of inhibitory input spikes. Therefore the hard problem of late inhibitory spikes can even be solved with a threshold that remains {\em constant} throughout the processing, i.e., even  before $t_{\rm min}^{(n)}$ . In practice we found that we could work with a constant threshold even if we did not  implement the strict condition 
 on the slope parameter formulated in Eq. (\ref{eq-stricter}) but worked instead with $ \alpha^{(n)}=1$. 
 The strict condition in Eq. (\ref{eq-stricter}) can lead to very large slope parameters which we might want to avoid in hardware implementations.

\subsubsection{Mapping with a dynamical threshold}
In the proof sketch we assumed 
a constant threshold $\vartheta_i^{(n)}$ for all times $t>t_{\rm min}^{(n)}$ . However, we can reinterpret the slope factor as a dynamical threshold. To see this, we intergrate 
Eq. (\ref{voltage-main}) and write the threshold condition that dermines the firing time $t_i^{(n)}<t_{\rm max}^{(n)}$
in the form
\be
\vartheta_i^{(n)} = V_i^{(n)}(t_i^{(n)}) = \alpha_i^{(n)} [t_i^{(n)} - t_{\rm min}^{(n-1)}] + \sum_j J_{ij}^{(n)} \epsilon(t_i^{(n)}-t_j^{(n-1)} )
\ee
where we have suppressed the external input and $\epsilon(s)$ is the voltage response to an input spike arriving at $s=0$ \cite{Gerstner02}.
Using standard textbook arguments, the term 
$\alpha_i^{(n)}[t_i^{(n)} - t_{\rm min}^{(n-1)}]$ can be shifted to the left-hand side which gives rise to a 'dynamical threshold' 
\cite{Gerstner02} defined as
$ \vartheta_i^{(n)}(t)  = \vartheta_i^{(n)} - 
\alpha_i^{(n)}[t - t_{\rm min}^{(n-1)}]$
.
Thus, the mapping in the corollary is identical to a mapping where the slope factor vanishes, but each neuron has a dynamical threshold that decreases linearly with time.


\subsubsection{Mapping with identical weights in SNN and ANN}
Previous studies have proposed approximative mappings under the condition
$J_{ij}^{(n)} = 
\bar{w}_{ij}^{(n)}$ for all neurons in all layers.
A quick glance at 
Eq. \ref{eq-J-mapping} tells us that  a mapping 
with $J_{ij}^{(n)} = 
\bar{w}_{ij}^{(n)}$
becomes  exact under the condition of a neuron-specific slope parameter 
\be\label{alphachoice}
 {\alpha_i^{(n)}} = {1-\sum_j \bar{w}_{ij}^{(n)}} \, . 
 \ee
 Thus, in contrast to the mapping in the proof  sketch of the Theorem, the slope parameter
 ${\alpha_i^{(n)}}$  is no longer a free parameter but {\em must} be chosen according to
 Eq. ({\ref{alphachoice}}) if the aim is to have the same set of weights in ANN and SNN.
 Interestingly, under this condition,  the trajectory of all neurons have the same slope of value one for $t>t_{\rm min}^{(n)}$.

\subsubsection{Mapping with less than one spike per neuron}
 Even though our theory requires each neuron to spike exactly once, it is possible to have an alternative implementation where a given spiking neuron fires only when the corresponding ReLU is {\em active}. Instead of sending (costly) spikes of inactive neurons, it is sufficient to store the reference times $t^{(n)}_{\mathrm{max}}$ for all $n$.   The trick is to set the slope of all trajectories of neurons $i$ in layer $n$ to $\alpha_i^{(n)} + \sum_j J_{ij}^{(n)}$ 
as soon as the maximum spike time $t^{(n-1)}_{\mathrm{max}}$ of neurons in layer $n-1$ has been reached. This is mathematically equivalent to making all inactive neurons in layer $n-1$ fire at time $t^{(n-1)}_{\mathrm{max}}$ but reduces the overall number of spikes in the network significantly. 
Therefore each neuron fires {\em at most} one spike. Since the ANN implements a nonlinear function from input to output, at least one neuron has to be inactive for at least one input data point, so that we know that on average there is strictly less than one spike per neuron.
Since we are interested in a low-energy solutions, we report in the following the  average number of active neurons  across all inputs and all neurons for the given dataset. This number can be interpreted as 'spikes per neuron per classification'. Note that this number depends on the specific regularization used during training of the ANN and can be further reduced by appropriate loss functions (which is out of scope of the present paper).

\subsection{Performance on Benchmark Datasets}

\par The above algorithm is a constructive proof that an exact mapping is possible. However, it is not clear how well it would perform in practice since there might be stability issues in the implementation or long processing delays that would reduce the attractivity of the mapping.  In the following we test  this algorithm  on several image classification tasks with different standard datasets. 

\par For each data set,  we report the classification accuracy for the original ReLU network, for the SNN,  as well as the percentage of agreement on a image-by-image basis between class prediction of original ReLU network and SNN network. Agreement is 100 percent, if for each image that is correctly (wrongly) classified by the ReLU, the image is also correctly (wrongly) classified by the SNN.
Furthermore we report percentage of spike per neuron under the implementation scheme mentioned at the end of the previous subsection.

\begin{table*}[t]
\centering
\begin{tabular}{l | c| c |c c| c | c} 
 \hline
  
 Model \& dataset & Image size & Classes &  \multicolumn{2}{|c|}{Accuracy [\%]} & Agreement [\%] & Spikes [\%] \\ 
  &  &  &  ReLU & SNN &  \\ 
 \hline\hline
 Fully connected, MNIST \cite{Rueckauer18} &28 $\times$ 28 $\times$ 1& 10 & 98.50 & 98.35 & - & - \\ 
 \textbf{Fully connected, MNIST [ours]} &28 $\times$ 28 $\times$ 1& 10 &98.52 & \textbf{98.52} & 100 & 50.28 \\ 
 LeNet5, MNIST \cite{Rueckauer18} &28 $\times$ 28 $\times$ 1& 10 &98.96 & 98.57 & - & - \\ 
 \textbf{LeNet5, MNIST (Fig. \ref{fig:Fig1}a) [ours]} &28 $\times$ 28 $\times$ 1& 10 &99.03 & \textbf{99.03} & 100 & 50.18 \\ 
 VGG16, MNIST [ours]&28 $\times$ 28 $\times$ 1& 10 & 99.60 & 99.60 & 100  & 51.21 \\ 
 VGG16, Fashion-MNIST [ours]&28 $\times$ 28 $\times$ 1& 10 & 93.70 & 93.70 & 100 & 45.34\\ 
VGG16, CIFAR10 \cite{yan_temporal} &32 $\times$ 32 $\times$ 3& 10 & 92.55 & 92.48 & - & - \\ 
\textbf{VGG16, CIFAR10  (Fig. \ref{fig:Fig1}d) [ours]} &32 $\times$ 32 $\times$ 3& 10 & 93.59 & \textbf{93.59} & 100 & 38.38 \\ 
  \hline\hline
 \multicolumn{7}{c}{ \textbf{Large-scale tests}}\\
  \hline\hline
 VGG16, CIFAR100 [ours]&32 $\times$ 32 $\times$ 3 & 100 & 70.48 & 70.48 &  100 & 38.21 \\ 
 VGG16, Places365 [ours] &224 $\times$ 224 $\times$ 3& 365 & 52.69 & 52.69 & 100 & 53.72 \\ 
 VGG16, PASS [ours] &224 $\times$ 224 $\times$ 3& 1000 & N/A & N/A & 100 & 53.24  \\ 
 \hline
\end{tabular}
\caption{
\begin{small}
Comparing performance of original ReLU networks and SNNs. Agreement metric shows percentage of inputs for which original ReLU network and SNN network predict the same class. Spikes column reveals the average percentage of active neurons across all inputs and layers. Places 365 and PASS are used as substitutes for ImageNet which breaches data privacy \cite{imagenet_privacy}. Accuracy calculation is not applicable for PASS dataset, as it is unlabeled.
\end{small}
} 
\label{tab:table}
\end{table*}

\subsubsection{MNIST, Fashion-MNIST and CIFAR 10}
In order to compare our results with existing conversion approaches \cite{Rueckauer18,yan_temporal}, we include MNIST and Fashion-MNIST \cite{mnist, fmnist} as well as CIFAR 10 \cite{cifar10_100} in our evaluation. We consider 16-layer VGG16, 5-layer LeNet5 and 2-layer fully connected networks, see Table \ref{tab:table}. VGG16 contains max pooling, fully connected and convolutional layers together with zero padding and batch normalization applied after ReLU activation functions. For the MNIST dataset the SNN achieves the same 99.6\%  accuracy as the original ReLU network with 100\% agreement, whereas the number of active neurons is around 51\%. Similarly, for Fashion-MNIST there is a 100\% agreement between SNN and ReLU predictions with the accuracy of 93.7\% and around 45\% of active neurons. In \cite{Rueckauer18} the authors perform a conversion of a 2-layer fully connected network as well as a LeNet5 convolutional network such that the weights and biases of the SNN and ANN are identical. We reproduce the ANN results of those models and compare the performance of their SNN with the one obtained using our method. Our SNN surpasses the accuracy in \cite{Rueckauer18}, and has 100\% agreement between SNN and ANN with around 50\% active neurons. The original and the scaled LeNet5 network can be seen in Fig. \ref{fig:Fig1}a. For the preprocessing and mapping details refer to the Methods.

\begin{figure*}[!t]
\centering
\includegraphics{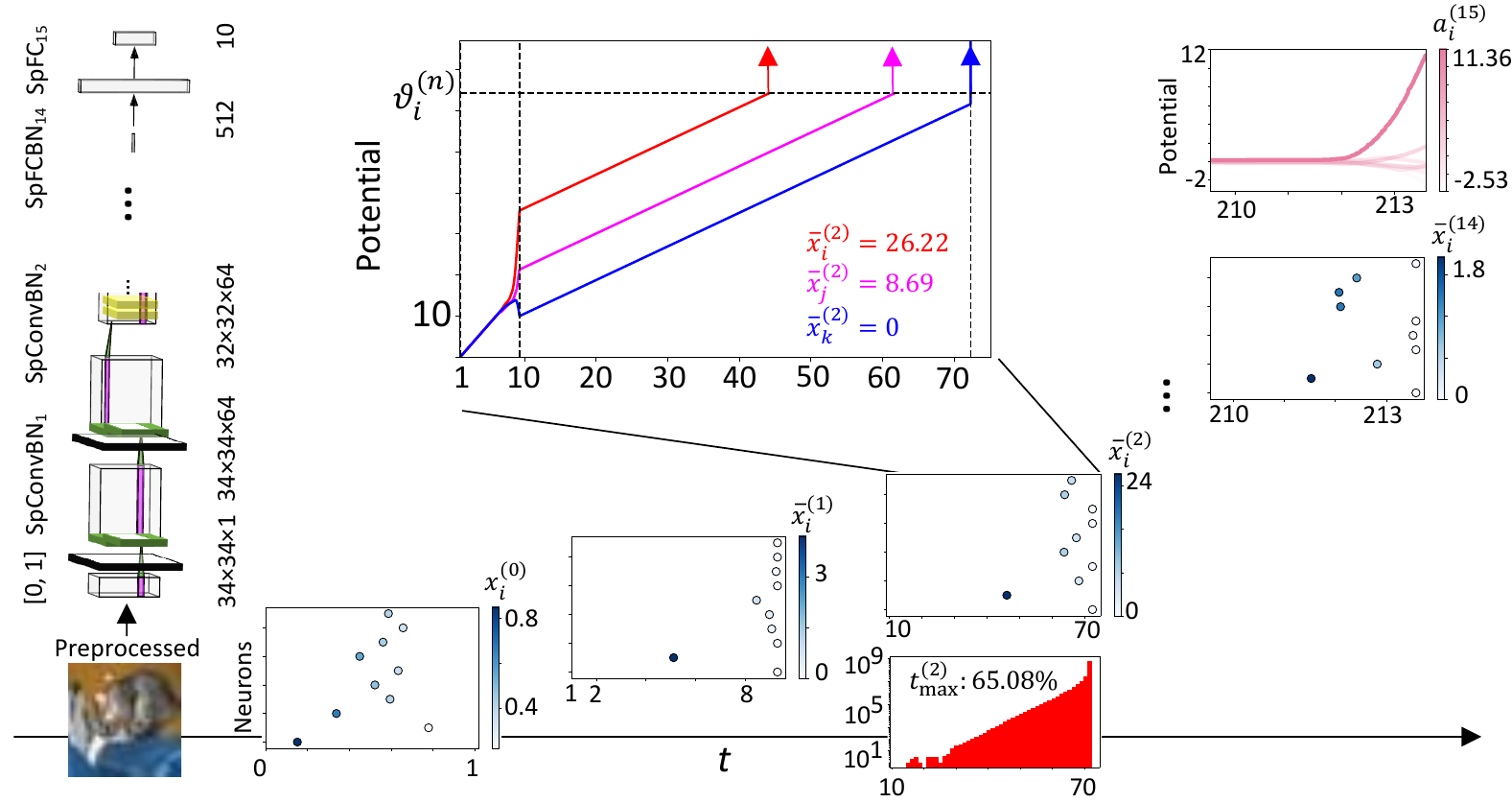}
\caption{
\begin{small}
{\bf Image Classification with spikes}. A presentation of a cat image from the CIFAR dataset  triggers spikes  (filled and open circles) in consecutive layers of the SNN. For layers 0, 1, 2 and 14, one of the neurons is the one that fires the earliest spike for this image whereas the other nine are randomly selected.  In a given layer, earlier spike times  correspond to larger values (darker color) of the corresponding ReLU  in the original network. At the output layer, the maximum potential corresponds to the largest activation variable in the ReLU network.
{\bf Zoom inset:}  Voltage trajectories of three neurons from the same convolutional channel in layer 2. Spike times correspond to the moments of threshold crossing. \textbf{Histogram inset:}  
Histogram of spike counts per time bin of neurons in layer 2 averaged across all neurons and all images in the test set. Only 34.92$\%$ of neurons fire before $t_{\rm max}^{(2)}$.
\end{small}
}
\label{fig:Fig3}
\end{figure*}


CIFAR10 contains color images of ten classes. The pretrained weights were obtained from an online repository \cite{cifar10_100} where a convolutional network similar to the VGG16 architecture (see Table \ref{tab:table}) was used. It comprises 15 layers in total since it uses only two fully connected layers instead of three (Fig. \ref{fig:Fig1}d). The network contains max pooling and convolutional layers together with zero padding and batch normalization applied after the ReLU activation functions. For CIFAR10 the SNN achieves the same 93.59\%  accuracy as the original ReLU network with 100\% agreement between the two networks, whereas at 38\% the number of active neurons is smaller than for the other datasets. Our scaled network is shown in Fig. \ref{fig:Fig1}d. 

\par In Fig. \ref{fig:Fig3} we show an example of an SNN inference for classification of a cat image from CIFAR10. For input and hidden layers a raster plot of 10 neurons is shown and the spikes of neurons with higher activation of the corresponding neuron in the ANN  are color-coded with darker shade. At the input layer the value of the data can be recovered from the spiking time of neuron $i$ as $x_i^0 = 1 - t_i^{(0)}$ and in layer $n$ the output of a neuron $i$ of scaled ReLU network can be recovered from the spiking time of the corresponding neuron in the SNN as $\bar{x}_i^{(n)} = t_{\mathrm{max}}^{(n)} - t_i^{(n)}$. The duration of the interval $[t_{\mathrm{min}}^{(n)},t_{\mathrm{max}}^{(n)}]$ varies considerably from one layer to the next. At the output layer $n=15$ a potential with darker color indicates a larger value of the activation variable of the corresponding neuron in the original ReLU network. At time $t^{(15)}_{\mathrm{max}}= t^{(14)}_{\mathrm{max}} +0.1$ when all the input from the layer $n=14$ has arrived, the maximum potential corresponds to the neuron with maximal activation variable, i.e. both networks predict the same class. 

\subsubsection{Large-scale data sets}
 We avoided the ImageNet dataset because of privacy-concerns \cite{Yang22} and used instead Places 365, PASS, and CIFAR 100 for more realistic tests.
The 'Places365-Standard' dataset contains high-resolution color images \cite{places365} resembling those in ImageNet dataset. The pretrained weights are obtained from an online repository \cite{places365_model} that contains a standard VGG16 network without batch normalization which we map to a corresponding SNN; see Table \ref{tab:table}. The SNN achieves the same 52.69\% accuracy as the original ReLU network with 100\% agreement between the two networks, whereas the number of active neurons is around 53\%. 

The PASS dataset consists of 1.4 million unlabeled images \cite{PASS} and is used as a substitute for ImageNet \cite{PASS} so as to avoid privacy-concerns. 
We use the same network as for the 'Places365-Standard' dataset, see Table \ref{tab:table}. The weights are downloaded from the VGG16 model for ImageNet available in TensorFlow \cite{tf_pretraned, imagenet}. An inference on an image from PASS returns one of the 1000 ImageNet classes as output. When performing inference with the SNN we verify the agreement of the class prediction between the two networks. There is a 100\% agreement between the original ReLU network and our SNN, with the fraction of active neurons around 53\%. 
The results of this and the previous paragraph together  show that the SNN  achieves the same accuracy as the corresponding ANN on ImageNet-like datasets using spiking neurons that fire  on average only for 53\% of the inputs. 

A similar statement is true for the CIFAR 100 dataset. Using the same network architecture as for CIFAR 10, and pretrained weights downloaded  from an online repository \cite{cifar10_100}, we find on CIFAR 100 a 100\% agreement between SNN and ReLU predictions with the accuracy of 70.48\% and around 38\% of active neurons.
Thus, on all tested large-scale datasets we find 100 percent agreement between the ANN and SNN indicating that the mapping is loss-free.

\begin{figure*}[!t]
\centering
\includegraphics{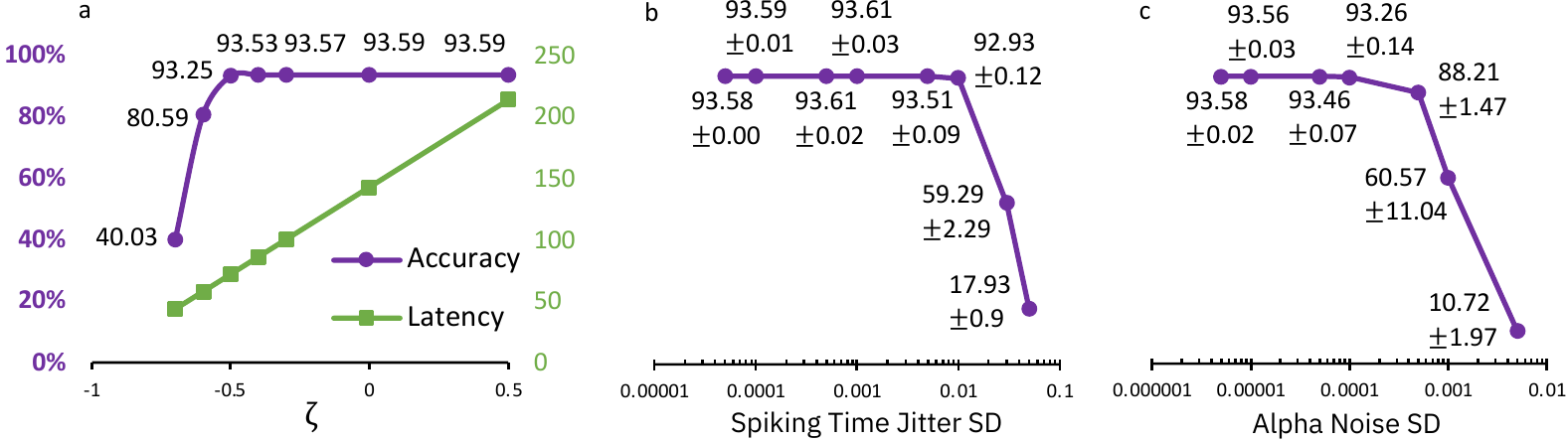}
\caption{
\begin{small}
{\bf Sensitivity tests} \textbf{a.} Performance as a function of the parameter $\zeta$. Note that the theorem requires $\zeta >0$; our standard choice in all other figures is $\zeta=0.5$. \textbf{b.} Performance of the network when there is a jitter in spike  timing; note that spike timing difference between the earliest and latest neuron in a given layer is often less than 2 time units (see Fig. \ref{fig:Fig3}). \textbf{c.} Performance of the networks when neuron-specific variations are added to the slope parameter $\alpha_i^{(n)}$ during inference; note that the reference value is $\alpha^{(n)}=1$ for all layers.
\end{small}
} 
\label{fig:Fig4}
\end{figure*}

\subsubsection{Sensitivity to noise and parameter changes}
As outlined in the introduction, the hardest problem of the conversion is to prevent spike firing in layer $n$ before all spikes from layer $n-1$ have arrived. In our mapping algorithm, a positive value $\zeta >0$ should guarantee, for a large enough and representative subsample of input images from the training set, that during test the above problem is avoided. For all implementation results so far, the standard choice was $\zeta=0.5$. In order to check sensitivity to the choice of $\zeta$, we varied $\zeta$ across positive and negative values. Using the 
 VGG16 model and the CIFAR10 dataset, we  found that the performance degrades gracefully when pushing $\zeta$ slightly into the negative regime, but breaks down for a value $\zeta < -0.5$, see Fig. \ref{fig:Fig4}a. Importantly, when switching from $\zeta=+0.5$ to $\zeta=-0.5$, the total processing time for image classification is reduced by a factor of three.

Noise in hardware implementations could potentially arise from a spike jitter caused for example by imprecisions in detecting the exact time of  threshold crossing. We add a Gaussian noise of given standard deviation (SD) and perform 16 trials. No performance degradation was observed up to a standard deviation of 0.001, see Fig. \ref{fig:Fig4}b.
  With a jitter of about 1 percent, the accuracy drops from 93.59\% to 92.93\%, which depending on the application may or may not be considered as acceptable.
  We note that spike times of hundred of neurons in a given  hidden layer spread over an interval of one or a few time units so that even with a jitter of 0.01 the order of spike firing is considerably changed. 

Imprecisions could also arise from heterogeneities in the hardware. A  sensitive parameter is the reference slope $\alpha^{(n)}$. We modify the slope parameter in a neuron-specific way  $\alpha_i^{(n)} + Y $ where $Y$ is a zero-mean Gaussian random variable with a standard deviation that we control. This simulates a systematic neuron-specific hardware manufacturing imperfection. Even a standard deviation of 0.001 leads to a dramatic drop in performance, see Fig. \ref{fig:Fig4}c. This is expected since a small mismatch in slope leads to a relatively large shift in spike timing because changes are accumulated throughout the integration interval $[t_{\mathrm{min}}^{(n-1)},t_{\mathrm{max}}^{(n)}]$.
As mentioned in the discussion, using existing learning rules for spiking neurons in the hardware loop \cite{Goltz21} could be used to rapidly fine-tuning weights to compensate for hardware heterogeneities.

\section {Discussion}
 In this paper we propose a specific mapping from a ReLU network to an SNN with time-to-first-spike coding that makes an SNN of integrate-and-fire units exactly equivalent to the deep ANN network. While a relation between ReLU networks and networks of non-leaky integrate-and-fire neurons has been suggested before \cite{Rueckauer18, Zhang21, Kheradpishe20}, there have been four obstacles that in the past prevented a successful exact mapping from deep artificial neural networks to deep spiking neural networks: 

\par (i) As mentioned in the introduction,  a neuron in layer $n$ that fires a spike before the last spike  from the neurons in the previous layer $n-1$  has arrived could compromise an exact mapping, since not all inputs are taken correctly into account: in particular, a late inhibitory input could have led to substantially different spiking time if taken into account. Having access to a representative sample of inputs from the training data enables us to solve this problem by an appropriate choice of intervals 
$[t^{(n)}_{\mathrm{min}},t^{(n)}_{\mathrm{max}}]$, with the condition
$t^{(n-1)}_{\mathrm{max}} = t^{(n)}_{\mathrm{min}}$. In other words, firing times of all neurons in layer $n$ fall into a desired interval, such that all spikes from layer $n-1$ have arrived before the first neuron in layer $n$ fires a spike. In practice, even for $\zeta=-0.5$, which significantly reduces the size of the interval,
the performance drops only slightly;
see Fig. \ref{fig:Fig4}a. This implies that the duration of the interval $[t^{(n)}_{\mathrm{min}},t^{(n)}_{\mathrm{max}}]$ can be considerably compressed compared to its theoretical value.
In the example of Fig. \ref{fig:Fig3} we observe that most neurons in the scaled ReLU network have an activity around 0, which implies that reducing the $\zeta$ parameter shifts only  the very few  most active neurons into a potentially problematic regime, but  does not influence the accuracy substantially.

\par (ii) In some implementations of an SNN, the slope of the potential of a neuron might be negative, zero, or only marginally  positive once all input spikes have arrived. In the last case, the threshold could be eventually reached but spiking would be sensitive to noise.
We have solved this problem by a positive slope parameter $\alpha_i^{(n)}$ for the trajectory of the  integrate-and-fire neuron in combination with a suitable (non-unique) preprocessing of ReLU parameters that together guarantee that the slope of the trajectory is larger than some minimal value once all input spikes have arrived. 

\par (iii) In the past it has been left open how to map the neuron of ReLU that is {\em inactive} for a given input vector to the corresponding spiking neuron. We have solved this problem by forcing the corresponding spiking neuron to fire a spike at the maximum spike time for that layer. We have also proposed an alternative implementation where inactive neurons do not fire spikes.

\par (iv) Existing conversion approaches often use custom activation functions or specific constraints during ANN training \cite{Rueckauer18, Neftci19, Zenke18,Zenke21, learning_gd_bio_bellec, learning_gd_snu, yan_temporal, rate_snn_bodo,  tdsnn_temporal,learning_gd_approx,learning_gd_prob}.
In contrast to prior work, our approach uses standard ANN elements and does not involve learning. The advantage of our approach in view of an application in  neuromorphic edge devices is that a  network consisting of standard fully connected and convolutional layers with ReLU activation function as well as max pooling and batch normalization can be pretrained using well-established optimization tools.  After conversion, the SNN is guaranteed to have the exact same accuracy as the original ANN.
The disadvantage is that hardware imperfections such as uncontrolled parameter variations are not taken into account during training.

TTFS coding for a conversion from ANN to SNN has been used before in an implementation 
\cite{Rueckauer18} that contains  elements similar to our approach, but with a few important differences. First, we have a systematic way to define the end $t_{\mathrm{max}}^{(n)}$ of the allowed spiking interval. Second, we use a TTFS code with a {\em linear} relation between spike times and ReLU output whereas the relation is nonlinear in the earlier scheme \cite{Rueckauer18}. Third, we identify for the case $\bar{w}_{ij}^{(n)} = J_{ij}^{(n)} $ an exact condition for the slope parameter and generalize to mappings where the weights are not simply copied from the ANN to the SNN. The latter gives the freedom to choose the slope parameter so that the trajectory has {\em always} positive slope.

\par The success of our method paves the road to many future research direction including both theory and application:

\par  (i) The discrete transition between spikes that are absent or present (depending on the input or on parameter variations) has plagued learning algorithms for spiking neural networks \cite{Goltz-Petrovici21, Kheradpishe20, Neftci19, Zenke18, Bohte2002, Tavanaei19}. Our theoretical contributions imply that spikes do not appear or disappear, but are rather shifted forward or backward within some finite interval. Earlier learning approaches have shown that those spikes that are triggered at moments when the slope of the  potential is close to zero induce a high sensitivity of spike timing to small parameter changes. By introducing a positive slope parameter into an integrate-and-fire neuron in combination with a suitable (non-unique) preprocessing of ReLU parameters our mapping guarantees that the slope of the trajectory is at the moment of firing bounded within some favourable range, so that the problems of sensitivity or discrete transitions are avoided. Therefore, our mapping approach opens the path towards stable learning algorithms in single-spike deep SNNs. 
\par (ii) Extension of the mapping to other architectures such as ResNet and to other types of neurons beyond non-leaky integrate-and-fire and ReLU would give the opportunity to have higher flexibility in terms of pretrained models. Moreover, it is of interest to further expand the theoretical framework such that it processes not just a single image but a stream of input data. This would present significant benefits for  applications.
\par  (iii)  To leverage our theoretical contribution for low-energy applications, a hardware implementation of this algorithm is desirable. In that context we are interested to further reduce the number of spikes and latency. With an improved implementation we have already reduced the number of spikes by roughly 50\%, see Table \ref{tab:table}. The latency can further be optimized by a less conservative choice of meta-parameters of the mapping so as to reduce the dead time between spike arrival times in layers $n-1$ and layer $n$. In particular a choice $\zeta=-0.5$ (instead of $\zeta=+0.5)$ reduces the overall processing time by a factor of three without a dramatic loss in performance; see Fig. \ref{fig:Fig4}a.

\par  (iv) For hardware implementations the question of robustness to noise and heterogeneities is also important. We have started to explore the robustness of our algorithm to noise by adding a Gaussian noise of given standard deviation to the spike times of each layer, see Fig. \ref{fig:Fig4}b. Moreover, we have considered the case where the conversion was done with slope parameter $\alpha^{(n)} = 1$, while the hardware introduces fixed noise of given standard deviation for the slope parameter of each neuron, see Fig. \ref{fig:Fig4}c.
It would be possible to fine-tune network weights with existing algorithm \cite{Goltz-Petrovici21} to compensate for hardware heterogeneities. In this context it would also be of interest to study the effects of weight quantization. Future work on this topic will eventually depend on the concrete hardware implementation that is envisaged.

To summarize,  this paper provides a constructive proof that deep ReLU networks and single-spike neural networks of integrate-and-fire neurons are equivalent.  As a consequence, we reach functional  deep spiking neural networks that have the same accuracy as ReLU networks and where spiking neurons fire at most one spike per neuron. Since spike transmission is a costly process in biology \cite{Attwell01} and neuromorphic hardware \cite{Sorbaro20}, our mathematical results open a pathway to low-energy computing with deep neural networks.


\section{Methods}

\subsection{Preprocessing}

Before we  perform the mapping from the ReLU network to the SNN, we perform a few preprocessing steps on the network with pretrained weights.

(i) If the network doesn't use batch normalization, this step is skipped. If batch normaliztion is implemented,  it is fused into the neighbouring fully connected and convolutional layers. The parameters of the batch normalization are $\hat{\mu_i}^{(n)}$ and $(\hat{\sigma_i}^{(n)})^2$ denoting the estimated mean and variance, $\gamma_i^{(n)}$ and $\beta_i^{(n)}$ which indicate scaling and shift factors learned during the optimization whereas $\epsilon$ is a small constant. In the following equations we use $\kappa_i^{(n)}$ to denote the scaling factor ${\gamma_i^{(n)}}/{\sqrt{(\hat{\sigma_i}^{(n)})^2+\epsilon}}$.   
\par When batch normalization is applied to the activation variable $a_i^{(n)}$ and before the activation function, it is fused with the processing of the previous layer (see Fig. \ref{fig:Fig1}a). The parameters are transformed as follows:

\be
b_{i}^{(n)} \leftarrow  \kappa_i^{(n)}(b_i^{(n)} - \hat{\mu_i}^{(n)}) + \beta_i^{(n)}
\ee

\be
\label{eq:bn_before_w}
w_{ij}^{(n)} \leftarrow \kappa_i^{(n)}w_{ij}^{(n)}.
\ee
Note that in case of convolutional architecture each index $i$ corresponds to a different channel.

\par When batch normalization is applied to the output of the activation function $x_i^{(n)}$, then it is fused with the processing of the subsequent layer (see Fig. \ref{fig:Fig1}d). The parameters are transformed as follows:
\be
\label{eq:bn_after_b}
b_{k}^{(n+1)} \leftarrow  b_{k}^{(n+1)}+ \sum_i(\beta_i^{(n)}-\kappa_i^{(n)}\hat{\mu_i}^{(n)}) w_{ki}^{(n+1)} ,
\ee

\be
\label{eq:bn_after_w}
w_{ki}^{(n+1)} \leftarrow \kappa_i^{(n)} w_{ki}^{(n+1)} . 
\ee
Note that the assignments of biases and weights need to be executed in this particular order. 
Moreover, in case of convolutional architecture, there are a few special cases  that need to be considered.
\par When batch normalization is applied to a zero-padded input into a convolutional layer, the bias change in Eq. (\ref{eq:bn_after_b}) introduces an unnecessary offset at zero-padded locations. For these particular locations, we calculate the bias by taking into account only the set of inputs $S_l$ which were not obtained through padding (see Fig. \ref{fig:Fig1}b). Eq. (\ref{eq:bn_after_b}) is replaced with:

\be
b_{k, l}^{(n+1)} \leftarrow b_{k, l}^{(n+1)} + \sum_{i\in S_l}(\beta_i^{(n)}-\kappa_i^{(n)}\hat{\mu_i}^{(n)}) w_{ki}^{(n+1)} 
\ee
\par When max pooling is applied after batch normalization, the weights of the subsequent convolutional layer are changed as described in Eqs. (\ref{eq:bn_after_b}) and (\ref{eq:bn_after_w}). The batch normalization  multiplies the output of each channel with factor $\kappa_i^{(n)}$, see Eq. (\ref{eq:bn_after_w}). When this value is negative, the sign of the output is changed. During the inference time, the max pooling operation is transformed into a min pooling operation for the channels with switched sign (see Fig. \ref{fig:Fig1}c).

\begin{algorithm}[t]
\caption{Preprocessing}
\label{alg:preproc}
\hspace*{\algorithmicindent} \textbf{Input:} Model with parameters $w_{ij}^{(n)}$ and $b_{i}^{(n)}$ \\
\hspace*{\algorithmicindent} \textbf{Output:} $\overline{\mathrm{Model}}$ with parameters $\bar{w}_{ij}^{(n)}$ and $\bar{b}_{i}^{(n)}$
\begin{algorithmic}[1]
\State  $x_i^{(0)} \gets \frac{x_i^{(0)}-p}{q-p}$ \Comment{\parbox[t]{.4\linewidth}}{Rescale inputs to [0, 1]}
\For{layer $\in$ Model} \Comment{\parbox[t]{.4\linewidth}}{Iterate over all layers in the Model}
    \If {layer $\in$ [ $\mathrm{Conv_1}$, $\mathrm{FC_1}$ ] and [p, q] $\neq$ [0,1]} 
    \State Model $\gets$ fuse\_BN\_after\_ReLU (Model, p, q) \Comment{\parbox[t]{.4\linewidth}}{Fuse imaginary batch normalization layer in case network was trained on [p, q] range}
    \ElsIf{layer = BN and layer+1 = ReLU} 
    \State Model $\gets$ fuse\_BN\_before\_ReLU (Model) \Comment{\parbox[t]{.4\linewidth}}{Fuse batch normalization before activation with previous parametrized layer} 
    \ElsIf{layer = BN and layer-1 = ReLU} 
    \State Model $\gets$ fuse\_BN\_after\_ReLU (Model) 
    \Comment{\parbox[t]{.4\linewidth}}{Fuse batch normalization after activation with next parametrized layer and process padding or max pooling, Figs. \ref{fig:Fig1}b, \ref{fig:Fig1}c} 
     \EndIf
\EndFor
\State $c_i^{(0)}$ $\gets$ 0, n $\gets$ 1 
\For{layer $\in$ Model}  \Comment{\parbox[t]{.4\linewidth}}{Iterate over all layers in the Model} 
 \If {layer $\in$ [ $\mathrm{Conv_n}$, $\mathrm{FC_n}$ ]}
    \State $ \overline{\mathrm{Model}}$, $c_i^{(n)} \gets$ 
     scale (Model, $c_i^{(n-1)}$, $B_{\mathrm{low}}$, $\delta$) \Comment{\parbox[t]{.4\linewidth}}{Scale layer with $c_i^{(n-1)}$, Eqs. (\ref{eq:scale_next_w_pos}), (\ref{eq:scale_next_w_neg}), and then with $c_i^{(n)}$, Eqs. (\ref{eq:scale_cur_w_pos}), (\ref{eq:scale_cur_b_pos}), (\ref{eq:scale_cur_w_neg}), (\ref{eq:scale_cur_b_neg})} 
     \State $X^{(n)} \gets$ max\_output ($\mu$, $\overline{\mathrm{Model}}$) \Comment{\parbox[t]{.4\linewidth}}{Calculate maximum output for layer n given training samples $\mu$} 
     \State n $\gets$ n + 1
  \EndIf
\EndFor  
\end{algorithmic}
\end{algorithm}

\par (ii) If network has input in range $[0, 1]$, this step is skipped. Let's assume that the network has input in arbitrary  range $[p, q]$. We would like for the network to operate for input in $[0, 1]$ interval without changing its output. This scaling can be seen as an imaginary batch normalization layer between the input layer and  the first layer. \par The input data is transformed as $x_i^{(0)} \leftarrow \frac{x_i^{(0)}-p}{q-p}$ and the biases and weights of the first layer are set to:
\be
\label{eq:input_scaling_b}
b_{k}^{(1)} \leftarrow  b_{k}^{(1)}+ p\sum_i w_{ki}^{(1)} ,
\ee

\be
\label{eq:input_scaling_w}
w_{ki}^{(1)} \leftarrow (q-p) w_{ki}^{(1)}  .
\ee

When there is zero padding in the first convolutional layer, Eq. (\ref{eq:input_scaling_b}) is replaced with:
\be
b_{k, l}^{(1)} \leftarrow b_{k, l}^{(1)} + p \sum_{i\in S_l}w_{ki}^{(1)}
\ee

\par (iii) In order to guarantee that the potential increases once all input spikes have arrived,  we rescale the parameters of the ReLU network. We exploit the scaling symmetry of ReLU neurons $[a_i]_+ = C [a_i/C]_+$, for $C>0$ and normalize weights so that the sum of input weights is smaller than $1-\delta$, for some $0 < \delta < 1$. 
Similarly, we want to make sure that the sum of input weights does not fall below some lower bound $(-B_{\mathrm{low}}) < 0$.
To implement the scaling, we begin from the initial weights $\bar{w}_{ij}^{(n)} \leftarrow {w}_{ij}^{(n)}$ and biases $\bar{b}_{i}^{(n)} \leftarrow {b}_{i}^{(n)}$, start in 
layer $n=1$ and proceed up to $n=M$ one layer at a time.
For each neuron $i$, we calculate the sum over all the incoming weights 
\be
c_i^{(n)}
= \sum_j \bar{w}_{ij}^{(n)} 
\ee
If
$c_i^{(n)}>(1-\delta)$, we set for this specific neuron $i$
 its incoming weights for all $j$ to
\be
\label{eq:scale_cur_w_pos}
\bar{w}_{ij}^{(n)} \leftarrow 
  \frac{(1-\delta)}{c_i^{(n)}}\bar{w}_{ij}^{(n)}
\ee
bias to
\be
\label{eq:scale_cur_b_pos}
\bar{b}_{i}^{(n)} \leftarrow 
  \frac{(1-\delta) }{c_i^{(n)}}\bar{b}_{i}^{(n)} 
\ee
and for all $k$ the outgoing weights to
\be
\label{eq:scale_next_w_pos}
    \bar{w}_{ki}^{(n+1)} \leftarrow \frac{c_i^{(n)}}{1-\delta}  \bar{w}_{ki}^{(n+1)}
\ee
Similarly, if
$c_i^{(n)}\le (-B_{\mathrm{low}})$ we set for
all $j$ the  incoming weights to 
\be
\label{eq:scale_cur_w_neg}
\bar{w}_{ij}^{(n)} \leftarrow  \frac{B_{\mathrm{low}}}{|c_i^{(n)}|}\bar{w}_{ij}^{(n)} 
\ee
bias to
\be
\label{eq:scale_cur_b_neg}
\bar{b}_{i}^{(n)} \leftarrow  \frac{B_{\mathrm{low}} }{|c_i^{(n)}|}\bar{b}_{i}^{(n)}
\ee
and for all $k$ the  outgoing weights to
    \be
    \label{eq:scale_next_w_neg}
    \bar{w}_{ki}^{(n+1)} \leftarrow \frac{|c_i^{(n)}| }{B_{\mathrm{low}}} \, \bar{w}_{ki}^{(n+1)}
    \ee
Note that signs are not changed by the scaling operation. Scaling ensures that for all hidden layers $(-B_{\mathrm{low}})\le \sum_j \bar{w}_{ij}^{(n)} < 1$. We have larger weights in the final output layer (readout weights), but this does not cause any problems. The network where all the above preprocessing steps are applied is called a \textit{scaled ReLU network}. Its parameters are denoted with a bar to distinguish them from the original, unscaled, network.

\par If $c_i^{(n)}>(1-\delta)$ the activation variable $\bar{a}_i^{(n)}$ is given by
 \be
 \label{eq:a_bar_pos}
    \bar{a}_i^{(n)} \leftarrow  
  \frac{(1-\delta)}{c_i^{(n)}}a_{i}^{(n)}
    \ee
and if $c_i^{(n)}\le (-B_{\mathrm{low}})$ by
 \be
  \label{eq:a_bar_neg}
     \bar{a}_i^{(n)} \leftarrow 
  \frac{B_{\mathrm{low}}}{|c_i^{(n)}|}a_{i}^{(n)}
    \ee
and $ \bar{a}_i^{(n)} \leftarrow a_i^{(n)}$ otherwise. Similarly, if $c_i^{(n)}>(1-\delta)$ the output $\bar{x}_i^{(n)}$ of ReLU is given by
 \be
  \label{eq:x_bar_pos}
    \bar{x}_i^{(n)} \leftarrow 
  \frac{(1-\delta) }{c_i^{(n)}}x_{i}^{(n)}
    \ee
and if $c_i^{(n)}\le (-B_{\mathrm{low}})$ as
 \be
  \label{eq:x_bar_neg}
     \bar{x}_i^{(n)} \leftarrow 
  \frac{B_{\mathrm{low}}}{|c_i^{(n)}|}x_{i}^{(n)}
    \ee
and $ \bar{x}_i^{(n)} \leftarrow x_i^{(n)}$ otherwise.

\par (iv) We apply all training data $1\le \mu \le P$ at the input layer of the scaled ReLU network and observe the activation pattern for each neuron in the network. For each layer $n$ we determine the maximal output of the activation function
$\bar{x}_i^{(n)}(\mu)$ across all training data $1\le \mu \le P$ and all neurons $i$ in that layer:
\be\label{eq-BigA}
X^{(n)}
= \max_{i,\mu} \{ \bar{x}_i^{(n)}(\mu)
\}
\ee
If the number $P$ is very large, we choose a statistically representative subset of data and perform the max-operation over these.

\begin{algorithm}[t]
\caption{Conversion}
\label{alg:mapping}
\hspace*{\algorithmicindent} \textbf{Input:} $\overline{\mathrm{Model}}$ with parameters $\bar{w}_{ij}^{(n)}$ and $\bar{b}_{i}^{(n)}$, $X^{(n)}$ \\
\hspace*{\algorithmicindent} \textbf{Output:} $\mathrm{SpModel}$ with parameters $J_{ij}^{(n)}$, $\vartheta_{i}^{(n)}$ $\alpha^{(n)}, t_{\mathrm{min}}^{(n)}$ and $t_{\mathrm{max}}^{(n)}$
\begin{algorithmic}[1]
\State $t_{min}^{(0)} \gets 0$, $t_{max}^{(0)} \gets 1$, $n \gets 1$, $\alpha^{(n)} \gets 1, \forall n \gets 1..M$  
\For{layer $\in \overline{\mathrm{Model}}$} \Comment{\parbox[t]{.4\linewidth}{Iterate layer-wise from the input to the output layer and calculate parameters and intervals}}
 \If {layer $\in$ [ $\mathrm{\overline{ConvBN}_n}$, $\mathrm{\overline{FCBN}_n}$ ]} 
    \If {layer+1=ReLU}
    \State  $t_{min}^{(n)} \gets t_{max}^{(n-1)}$
    \State  $t_{max}^{(n)} \gets t_{max}^{(n-1)} + (1+\zeta)X^{(n)}$ 
    \State $J_{ij}^{(n)} \gets \frac{\alpha^{(n)}\bar{w}_{ij}^{(n)}}{1 - \sum \bar{w}_{ij}^{(n)}}$
    \State $\vartheta_i^{(n)} \gets \alpha^{(n)}(t_{max}^{(n)}-t_{min}^{(n-1)}) + \sum_i J_{ij}^{(n)} (t_{max}^{(n)}-t_{min}^{(n)}) - (\alpha^{(n)}+\sum_i J_{ij}^{(n)})\bar{b}_i^{(n)}$
    \ElsIf {layer+1=softmax}
    \State $\alpha_i^{(n)} \gets \frac{\bar{b}_i^{(n)}}{(t_{max}^{(n-1)}-t_{min}^{(n-1)})}$
     \State $J_{ij}^{(n)} \gets \bar{w}_{ij}^{(n)}$
    \EndIf
    \State $n \gets n + 1$ 
  \EndIf
\EndFor  
\end{algorithmic}
\end{algorithm}

\subsection{Conversion to SNN}

The essential idea of the mapping from the ReLU neurons to the spiking neurons is that a positive activation leading to an output $\bar{x}_i^{(n)}=\bar{a}_i^{(n)}>0$
is identified with an {\em early}  firing time:
$t_i^{(n)}= t_{\mathrm{max}}^{(n)}
-\bar{a}_i^{(n)}$, 
whereas vanishing output $\bar{x}_i^{(n)}=0$ corresponds to firing at
$t_i^{(n)}= t_{\mathrm{max}}^{(n)}$.

The actual mapping is defined as follows (see Fig. \ref{fig:Fig2}). 

(i) Input encoding.
The input data lies in the interval $0 \le x_i^{(0)}< 1$ and we set $t_{\mathrm{min}}^{(0)} =0$, $t_{\mathrm{max}}^{(0)} =1$
and
$t_i^{(0)} =1 - x_i^{(0)}$.
With the parameters of the input layer fixed, we now proceed layer by layer from $n=1$ to $n=M$

\par (ii) We set $t_{\mathrm{min}}^{(n)} =
t_{\mathrm{max}}^{(n-1)}$ 
\par (iii) We set $t_{\mathrm{max}}^{(n)} =
t_{\mathrm{min}}^{(n)} + B^{(n)}$
with $ B^{(n)} = (1 + \zeta)\, X^{(n)}$
and  $\zeta >0$. 
The idea is that even the neuron with the strongest input must fire within the desired interval $[t_{\mathrm{min}}^{(n)},t_{\mathrm{max}}^{(n)}]$, i.e., not too early. 
Under the assumption that the test data comes from the same statistical distribution as the training data, a small value $\zeta \ll 1$ should in practice provide a sufficient safety margin. Indeed, if the training data set is large enough to be statistically representative, the probability that test data contains a point causing activation larger than $(1 + \zeta)\, X^{(n)}$ decreases rapidly with $\zeta$.
 The  range $[t_{\mathrm{min}}^{(n)},t_{\mathrm{max}}^{(n)}]$ is therefore large enough to encode all the values from layer $n$ of the rescaled ReLU network. 

\par (iv) For a given $\alpha_i^{(n)} > 0$ we first choose a  reference threshold $\tilde{\vartheta}_i^{(n)}$ in layer $n$ such that an integrator without any spike input would fire at $t_{\mathrm{max}}^{(n)}$. Hence
 for $t>t_{\mathrm{min}}^{(n)}$ the reference threshold is
\be\label{eq-threshold}
\tilde{\vartheta}_i^{(n)}
= \alpha_i^{(n)} \, [t_{\mathrm{max}}^{(n)}
  - t_{\mathrm{min}}^{(n-1)} ]
\ee
For the formal proof of the exact mapping, we set the reference threshold for 
$t\le t_{\mathrm{min}}^{(n)}$ to  a sufficiently high value $\tilde{\vartheta}_i^{(n)}=\Theta\to \infty$ for all times $t\le t_{\mathrm{min}}^{(n)}$. This ensures that no neuron in layer $n$ fires before $t_{\mathrm{min}}^{(n)}$. The value from Eq. (\ref{eq-threshold}) is used only for $t > t_{\mathrm{min}}^{(n)}$. 
However, for our practical algorithmic implementations 
we use the threshold  given in Eq. (\ref{eq-threshold}) throughout for all $t$.

\par (v) The actual threshold also depends on the bias and weights of the neuron.  To account for this, we  set the actual threshold of neuron $i$ in layer $n$ to a value
\be
\label{eq:thr_full}
\vartheta_i^{(n)} = \tilde{\vartheta}_i^{(n)} + D_i^{(n)}
\ee

With these parameter choices, an exact  mapping 
from ReLU network to an SNN is possible with a value
\be\label{eq-bias}
D_i^{(n)}
= [B^{(n)} \sum_j
  J_{ij}^{(n)}]
- [\alpha_i^{(n)}
  + \sum_j J_{ij}^{(n)} ]\,
\bar{b}_i^{(n)}.
\ee
and
weights 
\be\label{eq-J}
J_{ij}^{(n)} = \frac{\alpha_i^{(n)}}{1-\sum_j \bar{w}_{ij}^{(n)}} \,
\bar{w}_{ij}^{(n)}
\ee
where $\bar{w}_{ij}^{(n)}$ are the weights of the scaled ReLU network. Note that the denominator of Eq. (\ref{eq-J})
is always positive since $\sum_j \bar{w}_{ij}^{(n)}<1$.
  Hence the mapping does not change the sign of the weights.
The  inverse weight transform from SNN to ReLU is
\be\label{eq-w}
\bar{w}_{ij}^{(n)} = \frac{1}{\alpha_i^{(n)}+\sum_j {J}_{ij}^{(n)}}    {J}_{ij}^{(n)}
\ee
If $\sum_j \bar{w}_{ij}^{(n)}<1$ then the denominator in Eq. (\ref{eq-J}) is positive. This completes the conversion.

Note that we kept biases $\bar{b}_i^{(n)}$ as explicit parameters. However, following standard practice in the ANN literature, we could replace biases by an additional input neuron with connection weight equal to  $\bar{b}_i^{(n)}$. The equations above as well as those for weight rescaling in Phase 1 are to be used analogously in that case.

\vspace{5mm}

{\bf Lemma.} With the conversion rules Eqs. 
(\ref{eq-threshold}) to  (\ref{eq-J}) spike firing occurs at a value $ \bar{x}_i^{(n)} = t_{\mathrm{max}}^{(n)} - t_i^{(n)}$. 

{\em Proof. }
Let us integrate the 
differential equation (\ref{voltage-main}) of the integrate-and-fire units
which yields for $t_{\mathrm{min}}^{(n)}
< t < t_{\mathrm{max}}^{(n)}$ a voltage
\be\label{integrated}
V_i^{(n)}(t)
= [t-t_{\mathrm{min}}^{(n-1)}]\,
\alpha_i^{(n)} + \sum_j J_{ij}^{(n)} [t-t_j^{(n-1)}]\,,
\ee
where all neurons in layer $n-1$ have firing times
$t_j^{(n-1)} \le t_{\mathrm{max}}^{(n-1)} = t_{\mathrm{min}}^{(n)}$.
The firing time
$t_i^{(n)}$
of neuron $i$ in layer $n$ is given by the threshold condition 
$V_i^{(n)}(t_i^{(n)}) = \vartheta_i^{(n)}$. We exploit that neurons in layer $n-1$ that have not yet fired are forced to fire at $t_{\mathrm{max}}^{(n-1)}$.
We now insert the claims
$t_i^{(n)} = t_{\mathrm{max}}^{(n)} - \bar{x}_i^{(n)}$and $t_j^{(n-1)} = t_{\mathrm{max}}^{(n-1)} - \bar{x}_j^{(n-1)}$ into  Eq. (\ref{integrated}) and use Eqs. (\ref{eq-threshold}), (\ref{eq:thr_full}),(\ref{eq-bias}) as well as $t_{\mathrm{min}}^{(n)} =
t_{\mathrm{max}}^{(n-1)}$ and $B^{(n)}=t_{\mathrm{max}}^{(n)}-
t_{\mathrm{min}}^{(n)}$ to find
\be
\bar{x}_i^{(n)}
= \frac{1}{\alpha_i^{(n)} + \sum_j J_{ij}^{(n)} }
\sum_j J_{ij}^{(n)} \bar{x}_j^{(n-1)} + \bar{b}_i^{(n)} 
\ee
The Eq. (\ref{eq-w}) for the weights follows from a comparison of this formula with the ReLU equation $\bar{x}_i^{(n)} = \sum_j \bar{w}_{ij} \bar{x}_j^{(n-1)} + \bar{b}_i^{(n)}$; see
Eq. (\ref{eq-ReLU}) with $\bar{x}_i^{(n)} = [\bar{a}_i^{(n)}]_+$.
The solution is unique since trajectories have positive slope so that the threshold is reached at most once.

\subsection{Conversion of Max pooling}
If the ReLU network contains max pooling layers, the SNN contains layers performing max pooling and min pooling, outputting the earliest and latest spiking time respectively.  This functionality can be implemented with integrate-and-fire neurons such that each neuron fires exactly one spike. To this end we introduce  connections $K_{ij}^{(n-1)}$ within a given layer. A spike at time $t_j^{(n-1)}$ of a ReLU neuron $j$ in layer $n-1$ generates a pulse current, modeled by a Dirac delta pulse of total charge  $K_{ij}^{(n-1)}$, which is injected into neuron $i$ of the max pooling or min pooling operation belonging to the layer $n-1$. The voltage of neuron $i$ evolves according to 
\be
\frac{\mathrm{d}V_{i(\mathrm{MMP})}^{(n-1)}}{\mathrm{d}t} =  \sum_j K_{ij}^{(n-1)} \delta (t- t_j^{(n-1)}) 
\ee
If $V_{i(\mathrm{MMP})}^{(n-1)}$ crosses the threshold $\vartheta_{i(\mathrm{MMP})}^{(n-1)}$ at time $t$ then $t=t_{i(\mathrm{MMP})}^{(n-1)}$ is the firing time of neuron $i$.
For the layers which are preceded by a max pooling or min pooling operation the Eq. (\ref{voltage-main}) is replaced with:
\be\label{voltage}
\frac{\mathrm{d}V_i^{(n)}}{\text{d}t} = \alpha_i^{(n)}\, H(t-t_{\mathrm{min}}^{(n-1)})
  + \sum_j
  J_{ij}^{(n)}
  H (t- t_{j(\mathrm{MMP})}^{(n-1)} ) + I_i^{(n)}(t)
\ee
In case of the max pooling operation, all weights $K_{ij}^{(n)}$ are set to slightly larger values than the threshold value $\vartheta_{i(\mathrm{MMP})}^{(n)}$, such that the very first input spike triggers firing. In case of min pooling operation, parameters $K_{ij}^{(n)}$ are set to the value of $\vartheta_{i(\mathrm{MMP})}^{(n)}/Q < K_{ij}^{(n)} < \vartheta_{i(\mathrm{MMP})}^{(n)}/(Q-1)  $ where $Q$ is the total number of inputs. As a consequence, the very last input spike triggers the firing.

\subsection{Mapping of the output layer}
The output layer of the scaled ReLU network has a softmax  activation function and parameters $\{\bar{w}_{ij}^{(M+1)}, \bar{b}_{i}^{(M+1)}\} $. In the SNN we implement the output layer with an integrator unit, i.e. the neurons just integrate the currents and do not spike. A spike arriving at the output layer at time $t_j^{(M)}$ from a neuron in layer $M$ generates a step current input with amplitude $\bar{w}_{ij}^{(M+1)}$ into neuron $i$ of layer $M+1$. The voltage of neuron $i$ in layer $M+1$ evolves according to
\be\label{voltage-final-layer}
\frac{\text{d}V_i^{(M+1)}}{\text{d}t} = \alpha_i^{(M+1)}\, H(t-t_{\mathrm{min}}^{(M)})
  + \sum_j \bar{w}_{ij}^{(M+1)} H (t- t_j^{(M)} ) 
\ee
where $H$ denotes the Heaviside step function. The non-leaky integration starts at time $t_{\mathrm{min}}^{(M)}$ and lasts until time $t_{\mathrm{max}}^{(M)}$ and $\alpha_i^{(M+1)}$ takes value:
\be
\alpha_i^{(M+1)} = \frac{\bar{b}_i^{(M+1)}}{t_{\mathrm{max}}^{(M)} - t_{\mathrm{min}}^{(M)}}
\ee
The largest potential $V_i^{(M+1)}$ at time $t_{\mathrm{max}}^{(M)}$ determines the prediction.

\subsection{Final remarks regarding the mapping}

First, as mentioned in the results section, other mappings are also possible. For efficient coding with short latency, the aim is to choose parameters such that the resulting time intervals $[t_{\mathrm{min}}^{(n)},t_{\mathrm{max}}^{(n)}]$  are not too large, however large enough to encode all values from the ReLU network with sufficient temporal resolution and such that the firing times of different layers do not overlap. Note that (in contrast to leaky integration with time-constant $\tau$) a non-leaky integrator has no intrinsic time scale.
Second, it would be possible to start the integration of all integrate-and-fire units across all layers $n$ synchronously at time $t=0$, if we increase at the same time the threshold in layer $n$ by an amount $\alpha^{(n)}\,t_{\mathrm{min}}^{(n)}$.

\par Third, since ${\alpha_i^{(n)} + \sum_j J_{ij}^{(n)} }>0$
and all neurons in layer $n-1$ have fired at or before $t_{\mathrm{max}}^{(n-1)}$, the voltage trajectories $V_{i}^{(n)}$ of all neurons $i$ in layer $n$ have for $t>t_{\mathrm{max}}^{(n-1)}=t_{\mathrm{min}}^{(n)}$ a positive slope; see. Eq. (\ref{voltage-main}). 
If, after preprocessing, $\sum_j \bar{w}_{ij}^{(n)} \le 1-\delta$, then the maximal  slope of the trajectory at threshold is ${\alpha^{(n)}}/{\delta} $. Similarly, if after preprocessing  $\sum_j \bar{w}_{ij}^{(n)} \ge -B_{\mathrm{low}}$, then the minimal slope at the moment of firing is
${\alpha^{(n)}}/(1+ B_{\mathrm{low}})$. A small slope of the potential close to the threshold  should be avoided, since this increases the sensitivity to noise (in particular in view of combining with learning algorithms or unknown heterogeneities in the exact value of the slope). In practice, a value of $B_{\mathrm{low}}=10$ worked fine for our numerical simulations.


Fourth, we used a value of $\zeta=0.5$. If the training set is large and if we have access to all data in the training set, a positive but small $\zeta\to 0$ would be sufficient to guarantee that a neuron cannot fire 'too early'. However, the test set could potentially include data where the total activation is slightly larger than the maximal activation in the training set.
Since training set and test set arise, in principle, from the same statistical distribution, a parameter choice
$\zeta=0.5$ should provide a sufficient safety margin and this is confirmed in our simulations in the Results section.

\subsection{Datasets}
We consider six datasets of different sizes and complexity:
\par (i) MNIST and Fashion MNIST datasets contain greyscale images of size $28 \times 28$ which are labeled into ten classes. For each of the two datasets there are 60000 training images and 10000 testing images. Data preprocessing step includes normalizing pixel values to the $[0, 1]$ range and in the case of a fully connected network the input is also reshaped. The pretrained parameters of the original ReLU networks are obtained by training with backpropagation using Adam optimizer \cite{adam} with exponential learning rate schedule and standard cross-entropy loss. We apply dropout for regularization. In case of the VGG16 architecture the kernel was always of size 3 and the input of each convolutional operation is zero padded such that the shape at the output remains the same. Due to small input size, the first max pooling operation in the standard VGG16 architecture is omitted. The output of the convolutional part of VGG16 is of size 512 which is followed by two fully connected layers each containing 512 neurons and the output layer. The LeNet5 architecture has three convolutional, two max pooling and two fully connected layers with 84 and 10 neurons, see Fig \ref{fig:Fig1}a. Finally, the 2-layer fully connected network has one hidden layer with 600 units. LeNet5 and VGG16-like networks also contain batch normalization before and after ReLU function, respectively. 
\par In Fig. \ref{fig:Fig1}a we see the scaled LeNet5 network where the batch normalization is fused with previous convolutional and fully connected layers and the parameters of the network are scaled. For VGG16 network the batch normalization is fused with next convolutional and fully connected layers. Moreover, in this case the shift which appears due to zero padding is counter balanced with bias change at certain locations, see Fig. \ref{fig:Fig1}b, and every time batch normalization appears before max pooling, the channels whose sign is changed are replaced with min pooling, see Fig. \ref{fig:Fig1}c. Since the model is trained on $[0, 1]$ range there is no need to fuse an imaginary batch normalization after the input. In order to obtain the scaled ReLU network the parameters of the network are scaled. Finding the maximum output $X^{(n)}$ of each layer on the subset of the training set finalizes the preprocessing step (see Fig. \ref{fig:Fig0}). In the following mapping phase the parameters of SNN are calculated. 

\par (ii) CIFAR10 and CIFAR100 contain images of size $32 \times 32 \times 3$ \cite{cifar}. For each of the two datasets there are 50000 training images and 10000 testing images. The data preprocessing step includes normalizing data with given fixed mean and standard deviation as given in \cite{cifar10_100}. The network was trained on the data rounded to $[-3, 3]$ range. In preparation for SNN mapping and inference, the input $x_i^{(0)}$ is further preprocessed as $x_i^{(0)} \leftarrow \frac{x_i^{(0)}+3}{6}$. The kernel is always of size 3 and the input of each convolutional operation is zero padded such that the shape at the output stays the same. The output of the convolutional part of the VGG16 architecture has size 512 which is followed by two fully connected layers with 512 and 10 neurons. 
\par During the preprocessing, the batch normalization is fused with the next convolutional and fully connected layers and bias is changed in locations where the input is coming from the zero padding. When necessary, the max pooling function is replaced with min pooling. Since the model is trained on $[-3, 3]$ range the imaginary batch normalization is fused with first convolutional layer and in locations where the input is generated by zero padding the bias is changed. In order to obtain the scaled ReLU network the parameters of the network are scaled. Finding the maximum output $X^{(n)}$ of each layer on the subset of training set finalizes the preprocessing step (see Fig. \ref{fig:Fig0}). In the following mapping phase the parameters of SNN are calculated.

\par (iii) The images in Places365-Standard dataset are labeled into 365 scene categories. There are 1.8 million training images, 36500 validation images and 328500 test images. Since the labels for the test set are not publicly available, we report the metrics on the validation set. Data preprocessing step includes centralizing data around a given fixed mean and reshaping it to the size of $224\times224\times3$ as described in \cite{places365_model}. The network is trained on the data which can be rounded to $[-200, 200]$ interval. In preparation for SNN mapping and inference the input $x_i^{(0)}$ is further preprocessed as $x_i^{(0)}\leftarrow \frac{x_i^{(0)}+200}{400}$. Since the model is trained on $[-200, 200]$ range the imaginary batch normalization is fused with first convolutional layer and in locations where the input is generated by zero padding the bias is changed. In order to obtain the scaled ReLU network the parameters of the network are scaled. Finding the maximum output $X^{(n)}$ of each layer on the subset of training set finalizes the preprocessing step (see Fig. \ref{fig:Fig0}). In the following mapping phase the parameters of SNN are calculated.

\par (iv) We randomly sample 100000 testing and 5000 training images from PASS dataset. Most of the images in the dataset are colored and the few ones that are not are dropped during data preprocessing. The images are reshaped to $224 \times 224 \times 3$ and preprocessed with the same function as ImageNet for VGG16, which includes centering each color channel around zero mean. Since the model is trained on $[-200, 200]$ range, in preparation for SNN mapping and inference, the input $x_i^{(0)}$ is further preprocessed as $x_i^{(0)} \leftarrow \frac{x_i^{(0)}+200}{400}$ situating the input on the $[0, 1]$ range. Moreover, the imaginary batch normalization is fused with the first convolutional layer and in locations where the input is generated by zero padding the bias is changed. In order to obtain the scaled ReLU network the parameters of the network are scaled. Finding the maximum output $X^{(n)}$ of each layer on the subset of training set finalizes the preprocessing step (see Fig. \ref{fig:Fig0}). In the following mapping phase the parameters of SNN are calculated.

\vspace{1mm}

{\bf Acknowledgements}. The research of W.G. and G.B. was supported by a Sinergia grant (No CRSII5 198612) of the Swiss National Science Foundation. 

\begin{spacing}{0.5}
\begin{small}
\bibliographystyle{unsrt}
\bibliography{references}

\begin{thebibliography}{10}

\bibitem{energy_server}
Emma Strubell, Ananya Ganesh, and Andrew McCallum.
\newblock Energy and policy considerations for deep learning in nlp.
\newblock {\em arXiv preprint arXiv:1906.02243}, 2019.

\bibitem{gpt3}
Tom Brown, Benjamin Mann, Nick Ryder, Melanie Subbiah, Jared~D Kaplan, Prafulla
  Dhariwal, Arvind Neelakantan, Pranav Shyam, Girish Sastry, Amanda Askell,
  et~al.
\newblock Language models are few-shot learners.
\newblock {\em Advances in neural information processing systems},
  33:1877--1901, 2020.

\bibitem{binarized}
Matthieu Courbariaux, Itay Hubara, Daniel Soudry, Ran El-Yaniv, and Yoshua
  Bengio.
\newblock Binarized neural networks: Training deep neural networks with weights
  and activations constrained to+ 1 or-1.
\newblock {\em arXiv preprint arXiv:1602.02830}, 2016.

\bibitem{mobilenets}
Andrew~G Howard, Menglong Zhu, Bo~Chen, Dmitry Kalenichenko, Weijun Wang,
  Tobias Weyand, Marco Andreetto, and Hartwig Adam.
\newblock Mobilenets: Efficient convolutional neural networks for mobile vision
  applications.
\newblock {\em arXiv preprint arXiv:1704.04861}, 2017.

\bibitem{efficientnet}
Mingxing Tan and Quoc Le.
\newblock Efficientnet: Rethinking model scaling for convolutional neural
  networks.
\newblock In {\em International conference on machine learning}, pages
  6105--6114. PMLR, 2019.

\bibitem{neuromorphic_in_memory1}
Geoffrey~W Burr, Robert~M Shelby, Abu Sebastian, Sangbum Kim, Seyoung Kim,
  Severin Sidler, Kumar Virwani, Masatoshi Ishii, Pritish Narayanan, Alessandro
  Fumarola, et~al.
\newblock Neuromorphic computing using non-volatile memory.
\newblock {\em Advances in Physics: X}, 2(1):89--124, 2017.

\bibitem{neuromorphic_in_memory2}
Abu Sebastian, Manuel Le~Gallo, Geoffrey~W Burr, Sangbum Kim, Matthew
  BrightSky, and Evangelos Eleftheriou.
\newblock Tutorial: Brain-inspired computing using phase-change memory devices.
\newblock {\em Journal of Applied Physics}, 124(11):111101, 2018.

\bibitem{Goltz-Petrovici21}
Julian G{\"o}ltz, Andreas Baumbach, Sebastian Billaudelle, AF~Kungl, Oliver
  Breitwieser, Karlheinz Meier, Johannes Schemmel, Laura Kriener, and Mihai~A
  Petrovici.
\newblock Fast and deep neuromorphic learning with first-spike coding.
\newblock In {\em Proceedings of the neuro-inspired computational elements
  workshop}, pages 1--3, 2020.

\bibitem{gallego2020event}
Guillermo Gallego, Tobi Delbr{\"u}ck, Garrick Orchard, Chiara Bartolozzi, Brian
  Taba, Andrea Censi, Stefan Leutenegger, Andrew~J Davison, J{\"o}rg Conradt,
  Kostas Daniilidis, et~al.
\newblock Event-based vision: A survey.
\newblock {\em IEEE transactions on pattern analysis and machine intelligence},
  44(1):154--180, 2020.

\bibitem{Goltz21}
J.~G{\"o}ltz, L.~Kriener, A.~Baumbach, S.~Billaudelle, O.~Breitweiser,
  B.~Cramer, D.~Dold, A.F. Kungl, W.~Senn, J.~Schemmel, K.~Meier, and M.A.
  Petrovici.
\newblock Fast and energy-efficient neuromorphic deep learning with first-spike
  times.
\newblock {\em Nature Machine Intelligence}, 3:823–835, 2021.

\bibitem{dl_edge_merge}
Xiaofei Wang, Yiwen Han, Victor~CM Leung, Dusit Niyato, Xueqiang Yan, and
  Xu~Chen.
\newblock Convergence of edge computing and deep learning: A comprehensive
  survey.
\newblock {\em IEEE Communications Surveys \& Tutorials}, 22(2):869--904, 2020.

\bibitem{google_nn}
Amirali Boroumand, Saugata Ghose, Berkin Akin, Ravi Narayanaswami, Geraldo~F
  Oliveira, Xiaoyu Ma, Eric Shiu, and Onur Mutlu.
\newblock Google neural network models for edge devices: Analyzing and
  mitigating machine learning inference bottlenecks.
\newblock In {\em 2021 30th International Conference on Parallel Architectures
  and Compilation Techniques (PACT)}, pages 159--172. IEEE, 2021.

\bibitem{ai_on_edge}
Ziheng Jiang, Tianqi Chen, and Mu~Li.
\newblock Efficient deep learning inference on edge devices.
\newblock {\em ACM SysML}, 2018.

\bibitem{Rieke96}
F.~Rieke, D.~Warland, R.~de~Ruyter~van Steveninck, and W.~Bialek.
\newblock {\em Spikes - Exploring the neural code}.
\newblock MIT Press, Cambridge, MA, 1996.

\bibitem{Gerstner02}
W.~Gerstner and W.~K. Kistler.
\newblock {\em Spiking Neuron Models: single neurons, populations, plasticity}.
\newblock Cambridge University Press, Cambridge UK, 2002.

\bibitem{Pillow08}
J.W. Pillow, J.~Shlens, L.~Paninski, A.~Sher, A.~M. Litke, E.~J. Chichilnisky,
  and E.P. Simoncelli.
\newblock Spatio-temporal correlations and visual signalling in a complete
  neuronal population.
\newblock {\em Nature}, 454:995--999, 2008.

\bibitem{temporal_retina}
Tim Gollisch and Markus Meister.
\newblock Rapid neural coding in the retina with relative spike latencies.
\newblock {\em science}, 319(5866):1108--1111, 2008.

\bibitem{temporal_tactile}
Roland~S Johansson and Ingvars Birznieks.
\newblock First spikes in ensembles of human tactile afferents code complex
  spatial fingertip events.
\newblock {\em Nature neuroscience}, 7(2):170--177, 2004.

\bibitem{temporal_owls}
M~Fabiana Kubke, Dino~P Massoglia, and Catherine~E Carr.
\newblock Developmental changes underlying the formation of the specialized
  time coding circuits in barn owls (tyto alba).
\newblock {\em Journal of Neuroscience}, 22(17):7671--7679, 2002.

\bibitem{Optican87}
L.~M. Optican and B.~J. Richmond.
\newblock Temporal encoding of two-dimensional patterns by single units in
  primate inferior temporal cortex. 3.~\protect{I}nformation theoretic
  analysis.
\newblock {\em J. Neurophysiol.}, 57:162--178, 1987.

\bibitem{Thorpe96}
S.~Thorpe, D.~Fize, and C.~Marlot.
\newblock Speed of processing in the human visual system.
\newblock {\em Nature}, 381:520--522, 1996.

\bibitem{Thorpe01}
S.~Thorpe, A.~Delorme, and R.~\protect{Van Rullen}.
\newblock Spike-based strategies for rapid processing.
\newblock {\em Neural Networks}, 14:715--725, 2001.

\bibitem{Hung05}
C.P. Hung, G.~Kreiman, T.~Poggio, and J.J. {DiCarlo}.
\newblock Fast readout of object identity from macaque inferior temporal
  cortex.
\newblock {\em Science}, 310:863 -- 866, 2005.

\bibitem{Yamins16}
D.L.K. Yamins and J.J. \protect{DiCarlo}.
\newblock Using goal-driven deep learning models to understand sensory cortex.
\newblock {\em Nat. Neurosci.}, 19:356--365, 2016.

\bibitem{Rueckauer18}
Bodo Rueckauer and Shih-Chii Liu.
\newblock Conversion of analog to spiking neural networks using sparse temporal
  coding.
\newblock In {\em 2018 IEEE international symposium on circuits and systems
  (ISCAS)}, pages 1--5. IEEE, 2018.

\bibitem{comsa}
Iulia~M Comsa, Krzysztof Potempa, Luca Versari, Thomas Fischbacher, Andrea
  Gesmundo, and Jyrki Alakuijala.
\newblock Temporal coding in spiking neural networks with alpha synaptic
  function.
\newblock In {\em ICASSP 2020-2020 IEEE International Conference on Acoustics,
  Speech and Signal Processing (ICASSP)}, pages 8529--8533. IEEE, 2020.

\bibitem{mostafa}
H.~{Mostafa}.
\newblock Supervised learning based on temporal coding in spiking neural
  networks.
\newblock {\em IEEE Transactions on Neural Networks and Learning Systems},
  29(7):3227--3235, 2018.

\bibitem{Zhang21}
Malu Zhang, Jiadong Wang, Jibin Wu, Ammar Belatreche, Burin Amornpaisannon,
  Zhixuan Zhang, Venkata Pavan~Kumar Miriyala, Hong Qu, Yansong Chua, Trevor~E
  Carlson, et~al.
\newblock Rectified linear postsynaptic potential function for backpropagation
  in deep spiking neural networks.
\newblock {\em IEEE Transactions on Neural Networks and Learning Systems},
  33(5):1947--1958, 2021.

\bibitem{Kheradpishe20}
Saeed~Reza Kheradpisheh and Timoth{\'e}e Masquelier.
\newblock Temporal backpropagation for spiking neural networks with one spike
  per neuron.
\newblock {\em International Journal of Neural Systems}, 30(06):2050027, 2020.

\bibitem{stanojevic}
Ana Stanojevic, Evangelos Eleftheriou, Giovanni Cherubini, Stanis{\l}aw
  Wo{\'z}niak, Angeliki Pantazi, and Wulfram Gerstner.
\newblock Approximating relu networks by single-spike computation.
\newblock In {\em 2022 IEEE International Conference on Image Processing
  (ICIP)}, pages 1901--1905. IEEE, 2022.

\bibitem{Bu22}
T.~Bu, W.~Fang, J.~Ding, P.~Dai, Z.~Yu, and T.~Huang.
\newblock Optimal ann-snn conversion for high- accuracy and ultra-low-latency
  spiking neural networks.
\newblock {\em ICLR}, 2022.

\bibitem{Stockl21}
C.~Stockl and W.~Maass.
\newblock Optimized spiking neurons can classify images with high accuracy
  through temporal coding with two spikes.
\newblock {\em Nat. Mach. Intell.}, 3:230--238, 2021.

\bibitem{Neftci19}
Emre~O Neftci, Hesham Mostafa, and Friedemann Zenke.
\newblock Surrogate gradient learning in spiking neural networks: Bringing the
  power of gradient-based optimization to spiking neural networks.
\newblock {\em IEEE Signal Processing Magazine}, 36(6):51--63, 2019.

\bibitem{Zenke18}
Friedemann Zenke and Surya Ganguli.
\newblock Superspike: Supervised learning in multilayer spiking neural
  networks.
\newblock {\em Neural computation}, 30(6):1514--1541, 2018.

\bibitem{Bohte2002}
Sander~M Bohte, Joost~N Kok, and Han La~Poutre.
\newblock Error-backpropagation in temporally encoded networks of spiking
  neurons.
\newblock {\em Neurocomputing}, 48(1-4):17--37, 2002.

\bibitem{Tavanaei19}
Amirhossein Tavanaei, Masoud Ghodrati, Saeed~Reza Kheradpisheh, Timoth{\'e}e
  Masquelier, and Anthony Maida.
\newblock Deep learning in spiking neural networks.
\newblock {\em Neural networks}, 111:47--63, 2019.

\bibitem{Zenke21}
F.~Zenke and T.P. Vogels.
\newblock The remarkable robustness of surrogate gradient learning for
  instilling complex function in spiking neural networks.
\newblock {\em Neural Computation}, page 899–925, 2021.

\bibitem{learning_gd_bio_bellec}
Guillaume Bellec, Franz Scherr, Anand Subramoney, Elias Hajek, Darjan Salaj,
  Robert Legenstein, and Wolfgang Maass.
\newblock A solution to the learning dilemma for recurrent networks of spiking
  neurons.
\newblock {\em Nature communications}, 11(1):1--15, 2020.

\bibitem{learning_gd_snu}
Stanis{\l}aw Wo{\'z}niak, Angeliki Pantazi, Thomas Bohnstingl, and Evangelos
  Eleftheriou.
\newblock Deep learning incorporating biologically inspired neural dynamics and
  in-memory computing.
\newblock {\em Nature Machine Intelligence}, 2(6):325--336, 2020.

\bibitem{yan_temporal}
Zhanglu Yan, Jun Zhou, and Weng-Fai Wong.
\newblock Near lossless transfer learning for spiking neural networks.
\newblock In {\em Proceedings of the AAAI Conference on Artificial
  Intelligence}, volume~35, pages 10577--10584, 2021.

\bibitem{cifar10_100}
Yonatan Geifman.
\newblock Github, 2018.

\bibitem{imagenet}
Olga Russakovsky, Jia Deng, Hao Su, Jonathan Krause, Sanjeev Satheesh, Sean Ma,
  Zhiheng Huang, Andrej Karpathy, Aditya Khosla, Michael Bernstein,
  Alexander~C. Berg, and Li~Fei-Fei.
\newblock {ImageNet Large Scale Visual Recognition Challenge}.
\newblock {\em International Journal of Computer Vision (IJCV)},
  115(3):211--252, 2015.

\bibitem{imagenet_privacy}
Kaiyu Yang, Jacqueline~H. Yau, Li~Fei-Fei, Jia Deng, and Olga Russakovsky.
\newblock A study of face obfuscation in {I}mage{N}et.
\newblock In Kamalika Chaudhuri, Stefanie Jegelka, Le~Song, Csaba Szepesvari,
  Gang Niu, and Sivan Sabato, editors, {\em Proceedings of the 39th
  International Conference on Machine Learning}, volume 162 of {\em Proceedings
  of Machine Learning Research}, pages 25313--25330. PMLR, 17--23 Jul 2022.

\bibitem{places365}
Bolei Zhou, Agata Lapedriza, Aditya Khosla, Aude Oliva, and Antonio Torralba.
\newblock Places: A 10 million image database for scene recognition.
\newblock {\em IEEE transactions on pattern analysis and machine intelligence},
  40(6):1452--1464, 2017.

\bibitem{PASS}
Yuki~M Asano, Christian Rupprecht, Andrew Zisserman, and Andrea Vedaldi.
\newblock Pass: An imagenet replacement for self-supervised pretraining without
  humans.
\newblock {\em arXiv preprint arXiv:2109.13228}, 2021.

\bibitem{mnist}
Li~Deng.
\newblock The mnist database of handwritten digit images for machine learning
  research.
\newblock {\em IEEE Signal Processing Magazine}, 29(6):141--142, 2012.

\bibitem{fmnist}
Han Xiao, Kashif Rasul, and Roland Vollgraf.
\newblock Fashion-mnist: a novel image dataset for benchmarking machine
  learning algorithms.
\newblock {\em arXiv preprint arXiv:1708.07747}, 2017.

\bibitem{Yang22}
K.~Yang, J.H. Yau, J.~Deng, and O.~Russakovsky.
\newblock A study of face obfuscation in imagenet.
\newblock {\em Proc. 39th Intern. Conf. Machine Learning PMLR},
  162(1):25313--25330, 2022.

\bibitem{places365_model}
Bolei Zhou.
\newblock Github, 2018.

\bibitem{tf_pretraned}
Applications - vgg16.
\newblock
  \url{https://www.tensorflow.org/api\_docs/python/tf/keras/applications/vgg16/VGG16}.
\newblock Accessed: 2022-11-21.

\bibitem{rate_snn_bodo}
Bodo Rueckauer, Iulia-Alexandra Lungu, Yuhuang Hu, Michael Pfeiffer, and
  Shih-Chii Liu.
\newblock Conversion of continuous-valued deep networks to efficient
  event-driven networks for image classification.
\newblock {\em Frontiers in neuroscience}, 11:682, 2017.

\bibitem{tdsnn_temporal}
Lei Zhang, Shengyuan Zhou, Tian Zhi, Zidong Du, and Yunji Chen.
\newblock Tdsnn: From deep neural networks to deep spike neural networks with
  temporal-coding.
\newblock In {\em Proceedings of the AAAI conference on artificial
  intelligence}, volume~33, pages 1319--1326, 2019.

\bibitem{learning_gd_approx}
Dongsung Huh and Terrence~J Sejnowski.
\newblock Gradient descent for spiking neural networks.
\newblock In S.~Bengio, H.~Wallach, H.~Larochelle, K.~Grauman, N.~Cesa-Bianchi,
  and R.~Garnett, editors, {\em Advances in Neural Information Processing
  Systems}, volume~31. Curran Associates, Inc., 2018.

\bibitem{learning_gd_prob}
Brian Gardner, Ioana Sporea, and Andr{\'e} Gr{\"u}ning.
\newblock Learning spatiotemporally encoded pattern transformations in
  structured spiking neural networks.
\newblock {\em Neural computation}, 27(12):2548--2586, 2015.

\bibitem{Attwell01}
D.~Attwell and S.B. Loughlin.
\newblock An energy budget for signaling in the grey matter of the brain.
\newblock {\em J. Cerebral Blood Flow and Metabolism}, 21:133–1145, 2001.

\bibitem{Sorbaro20}
M.~Sorbaro, Q.~Liu, M.~Bortone, and S.~Sheik.
\newblock Optimizing the energy consumption of spiking neural networks for
  neuromorphic applications.
\newblock {\em Front. Neurosci.}, 14:662, 2020.

\bibitem{adam}
Diederik~P Kingma and Jimmy Ba.
\newblock Adam: A method for stochastic optimization.
\newblock {\em arXiv preprint arXiv:1412.6980}, 2014.

\bibitem{cifar}
Alex Krizhevsky, Geoffrey Hinton, et~al.
\newblock Learning multiple layers of features from tiny images.
\newblock 2009.

\end{thebibliography}
\end{small}
\end{spacing}

\end{document}